\pdfoutput=1

\documentclass[11pt]{article}

\usepackage[table]{xcolor}

\usepackage[final]{acl}
\usepackage{graphicx}
\usepackage{times}
\usepackage{latexsym}

\usepackage[T1]{fontenc}

\usepackage[utf8]{inputenc}

\usepackage{microtype}
\usepackage{pdfpages}
\usepackage{inconsolata}
\usepackage{amssymb}
\usepackage{pifont}
\usepackage{multirow}
\usepackage{array}
\usepackage{float}
\usepackage{pdfpages}
\usepackage{algorithm}
\usepackage{algpseudocode}
\usepackage[most]{tcolorbox}

\def\codebaselink{https://kagnlp.github.io/codesim.github.io/}

\def\codebase{\href{\codebaselink}{\codebaselink}}

\newcommand{\tool}{\textsc{CodeSim}~}
\newcommand{\toolnospace}{\textsc{CodeSim}}

\def\ourapproach{\toolnospace}

\newenvironment{nobreakwords}
  {\sloppy\hyphenpenalty=10000\exhyphenpenalty=10000 }
  {\par}

\definecolor{BLUE}{RGB}{30,60,220} 
\definecolor{GREEN}{RGB}{2,100,51} 

\title{
\textsc{\textbf{
\textcolor{BLUE}{\textsc{Code}}\textcolor{GREEN}{\textsc{SIM}}}}: Multi-Agent Code Generation and Problem Solving through Simulation-Driven Planning and Debugging
}

\author{
Md. Ashraful Islam\thanks{Work done when working as a remote RA at QCRI.}$^{1}$, \ Mohammed Eunus Ali$^1$, \ Md Rizwan Parvez$^2$ \\
$^1$Bangladesh University of Engineering and Technology (BUET) \\
$^2$Qatar Computing Research Institute (QCRI) \\
\{mdashrafulpramanic, mohammed.eunus.ali\}@gmail.com, mparvez@hbku.edu.qa 
}

\begin{document}
\maketitle

\begin{abstract}
\begin{nobreakwords}
Large Language Models (LLMs) have made significant strides in code generation and problem solving. Current approaches employ external tool-based iterative debuggers that use  compiler or other tool-based runtime feedback to refine coarse programs generated by various methods. However, the effectiveness of these approaches heavily relies on the quality of the initial code generation, which remains an open challenge. In this paper, we introduce {\bf \toolnospace}, a novel multi-agent code generation framework that comprehensively addresses the stages of program synthesis—planning, coding, and debugging—through a human-like perception approach.  As human verifies their understanding of any algorithms through visual simulation, \tool uniquely features a method of plan verification and internal debugging through the step-by-step simulation of input/output. Extensive experiments across seven challenging competitive problem-solving and program synthesis benchmarks demonstrate \toolnospace's remarkable code generation capabilities. Our framework achieves new state-of-the-art (pass@1) results—({HumanEval \bf 95.1\%}, {MBPP \bf 90.7\%}, {APPS \bf 22\%}, and {CodeContests \bf 29.1\%}). Furthermore, our method shows potential for even greater enhancement when cascaded with external debuggers. To facilitate further research and development in this area, we have open-sourced our framework in this link (\codebase).
\end{nobreakwords}
\end{abstract}
\section{Introduction}
\label{sec:intro}

\begin{figure*}
    \centering
    \vspace{-4mm}
    \includegraphics[width=0.99\textwidth]{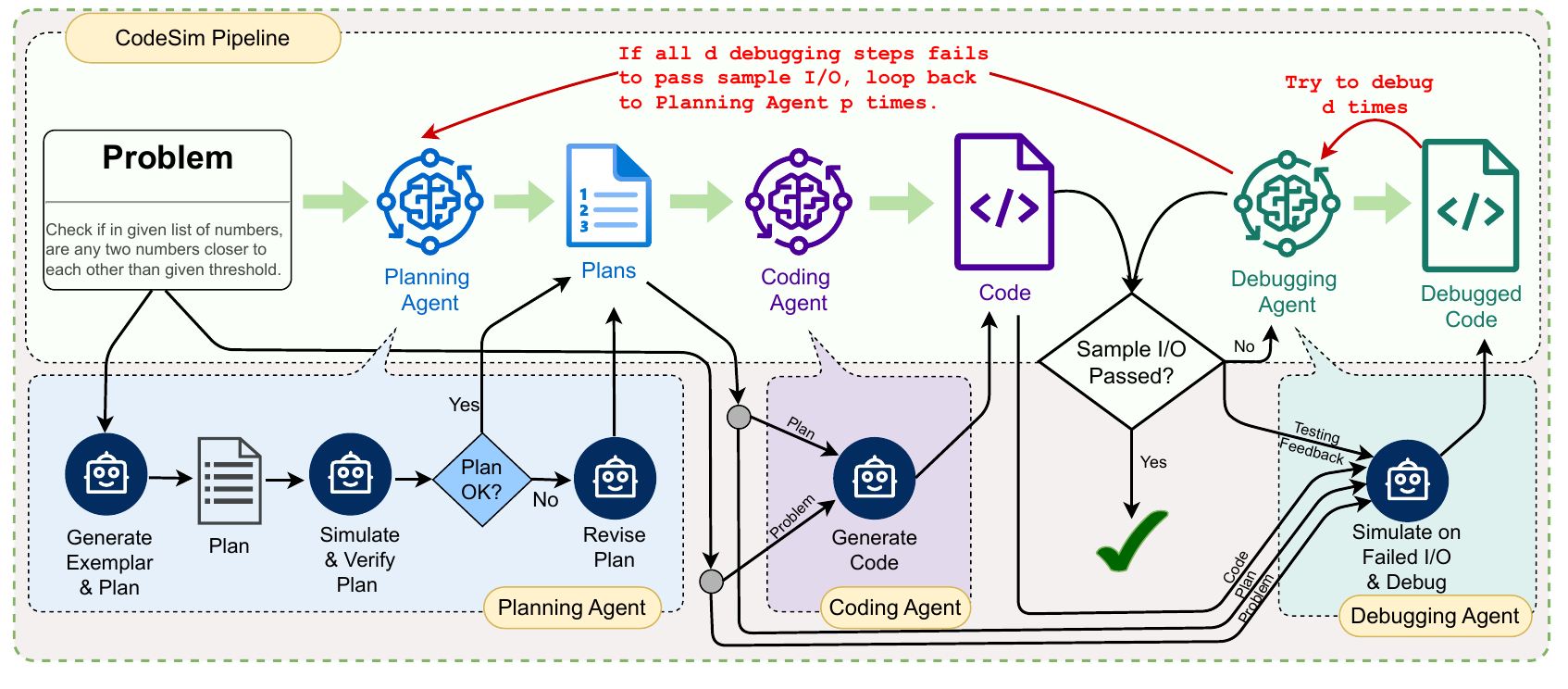}
    \vspace{-4mm}
    \caption{Overview of \toolnospace: It consists of three agents—planning, coding, and debugging. The \emph{Planning Agent} first generates an exemplar problem-solution (i.e., via self-retrieval) and devises a plan, which is then verified and refined through simulation. Next, the \emph{Coding Agent} implements the plan. Finally, the \emph{Debugging Agent} addresses potential bugs through step-wise simulation across $d$ trials. The entire process iterates $p$ times.
    }
    \label{fig:overview}
    \vspace{-4mm}
\end{figure*}

In recent years, the rise of Large Language Models (LLMs) has made significant advances in AI-assisted coding and reshaped the domain of code generation and problem-solving \citep{LLMSurvey}. Code generation assistants built on GPT-4 \citep{openai2024gpt4technicalreport}, Mistral \citep{jiang2023mistral}, and Llama \citep{dubey2024llama3herdmodels}, inter alia, have demonstrated unprecedented ability to understand, generate, and manipulate code across various programming languages and problem domains. However, despite these advancements, significant challenges persist in generating code for complex programming tasks. 


Current state-of-the-art approaches in code generation typically employ a \emph{dual-pass} process \cite{shi2024code-mgdebug, jin2024rgd, zhong-etal-2024-debug, levin2024chatdbg}. In the \emph{first pass}, they use LLMs to generate an initial,  fully/partially correct version of the program. Then accordingly in the \emph{second pass}, they apply external tool-based iterative debuggers that leverage runtime compiler feedback or other diagnostic tools to refine and correct the generated code. While this approach has shown promise, it necessitates numerous iterations of LLM-tool interactions, and importantly its effectiveness is heavily dependent on the quality of the initial code generation—a process that continues to present substantial difficulties. Therefore, in this paper, we present {\bf \toolnospace}, a novel multi-agent code generation framework that seamlessly synthesizes complex code solutions without external resources, while offering potential for further enhancement through minimal external debugging.

Synthesizing programs even in the \emph{first pass}, however, is fundamentally challenging, requiring a deep understanding of natural language processing, computer algorithms, data structures, and problem-solving strategies. These challenges are further compounded when attempting to generate code for competitive programming problems or advanced software engineering tasks, where adherence to specific constraints or passing unit tests are paramount \cite{khan2023xcodeeval}.

While earlier code generation methods employed direct approaches \cite{chen2021codex}, chain-of-thought reasoning \citep{CoT}, synthesized test-case guidance \cite{codet}, retrieval-augmented generation \cite{parvez2021retrieval}, and various in-context exemplar-based strategies \citep{shum-etal-2023-automatic, zhang2022automatic}, recent paradigms have shifted toward plan-based \cite{jiang2023self}, sampling or tree-searching \cite{zhou2023lats}, self-retrieval \cite{yasunaga2023large}, and diverse agent-based approaches \cite {zhang-etal-2024-codeagent, qian-etal-2024-chatdev, shinn2023reflexion, huang2023agentcoder, dong2023selfcollaboration}.


Most recently, MapCoder \cite{islam-etal-2024-mapcoder} proposes a multi-agent framework that implements agents emulating different stages of program synthesis such as recalling relevant examples, designing/planning, code generation, and testing/debugging. While this approach mimics a real developer's code generation cycle and shows improvements, it focuses solely on expanding steps without verifying the underlying hypotheses, with tests being performed only during the debugging phase. Consequently, the resulting gains are limited and it also requires larger number of iterations (i.e., LLM API calls).

\begin{figure*}[h]
    \centering
    \includegraphics[width=0.99\textwidth]{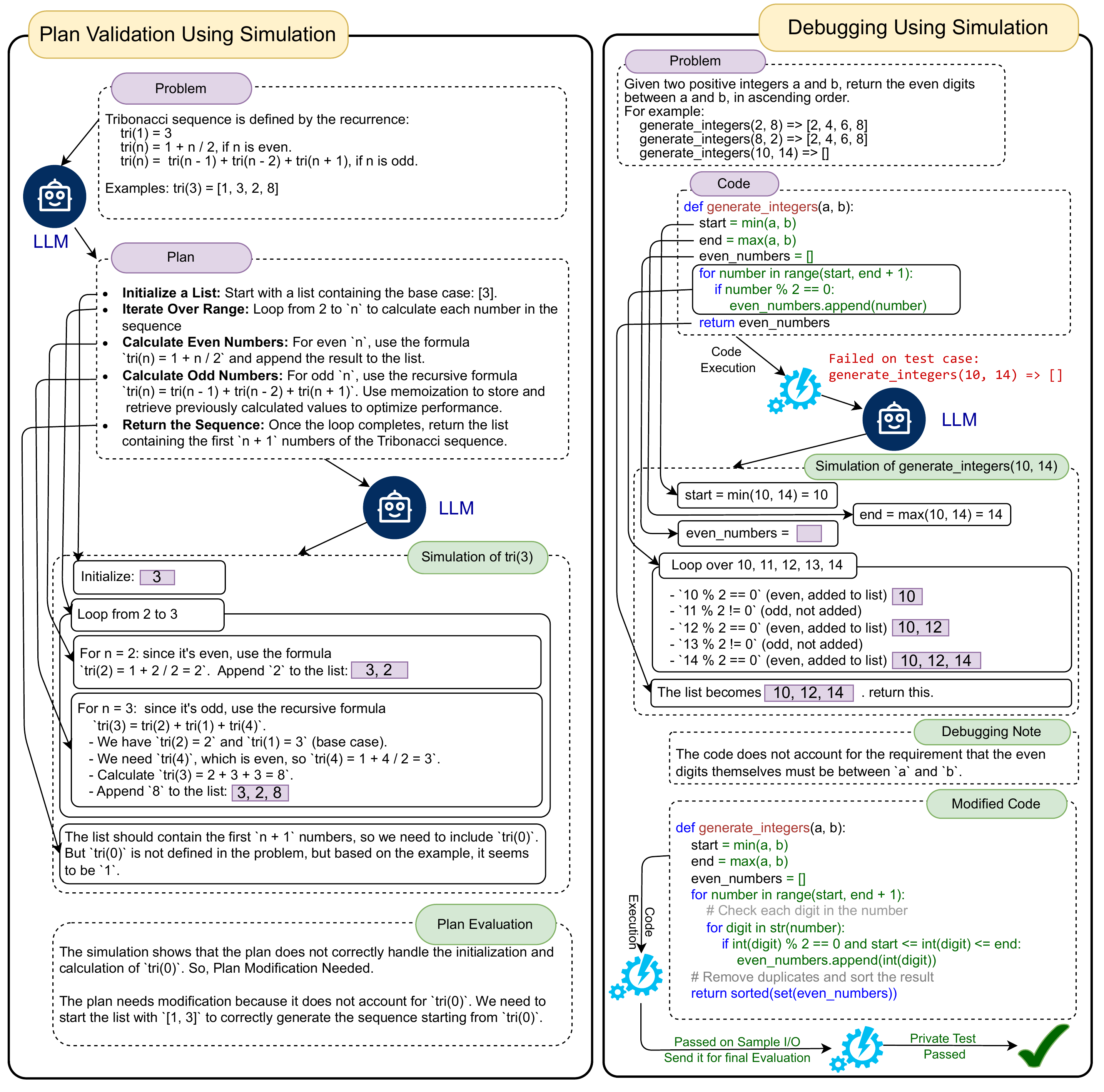}
    \caption{Example of Plan Validation using Simulation (left) and Debugging using Simulation (right) on two different problems using \toolnospace.}
    \label{fig:qualitative-example}
\end{figure*}

To address these limitations, \toolnospace—built upon planning, coding, and debugging agents—introduces a novel verification approach inspired by human problem-solving. By simulating input/output step-by-step, \tool verifies both the generated plans and performs internal debugging, mirroring how humans understand, visualize, and refine algorithms. This simulation-driven planning and debugging process ensures that each step is thoroughly evaluated, significantly enhancing both solution quality and efficiency. Figure \ref{fig:overview} shows an overview of our proposed approach, \tool and in Figure \ref{fig:qualitative-example}, we demonstrate how simulation assists in both plan verification and debugging, highlighting its crucial role in improving problem-solving accuracy.

We evaluate \tool on seven popular programming synthesis benchmarks, including foundational tasks like HumanEval and MBPP, as well as challenging competitive problem-solving benchmarks such as APPS, and CodeContest. Our experiments leverage multiple LLMs, including ChatGPT, GPT-4, GPT-4o, LLaMa, Gemma, and Mixtral, showcasing significant improvements in their program synthesis capabilities. \tool consistently achieves state-of-the-art performances, often surpassing strong baselines like MapCoder. Additionally, our findings suggest that \toolnospace's performance can be further improved when integrated with external debugging tools, such as LDB \cite{zhong-etal-2024-debug}, highlighting a promising direction for future research in hybrid code generation and debugging systems. Through detailed ablation studies, we provide valuable insights into \toolnospace's functionality. We will open-source our framework to support future research in AI-assisted programming and problem-solving.

\section{Related Work}
\label{sec:related-works}
\textbf{Code Generation:}
Program synthesis has been a fundamental challenge in AI for decades \cite{Zohar71}. Early attempts with smaller language models centered on code generation by fine-tuning neural networks \cite{wang2021codet5, ahmad2021unified, feng2020codebert, parvez-etal-2018-building, deep_net_for_source_code, code_gen_parsing, cmu_code_gen, naturalnessofsoft}, while others explored leveraging data flow information or conversational intents to guide the process \cite{andreas-etal-2020-task, yu-etal-2019-cosql}. Various prior approaches have also addressed code generation tasks using techniques such as data flow analysis and search-based methods \cite{li2022competition, parisotto2017neural, polozov2015flashmeta, gulwani2011automating}.
\smallskip \\
\noindent{\bf LLMs for Code:}
Various LLMs have been developed for code synthesis~\citep{austin2021program, chen2021evaluating, nijkamp2022codegen, fried2022incoder, allal2023santacoder, alphacode}. Recent open-source LLMs include the Llama family (Llama-2, CodeLlama, Llama3.1, etc.) \citep{roziere2023code, touvron2023llama}, the Mistral family (Mistral, Mixtral, Codestral) \citep{jiang2023mistral}, the Deepseek family (Deepseek Coder, Deepseek-V2, etc.) \cite{guo2024deepseek}, MoTCoder \cite{li2023motcoder}, and the Qwen family (Qwen 1.5, 2.5, 2.5-coder, etc.) \cite{hui2024qwen2}, all of which are capable of solving many basic problems.
\smallskip \\
\noindent{\bf Prompting LLMs and Multi-Agent Code Generation:}
LLM prompting can be summarized into three categories: retrieval \cite{yasunaga2023large, parvez2021retrieval, parvez-etal-2023-retrieval}, planning \citep{jiang2023self, wei2022chain}, and debugging \citep{le2022coderl, chen2022codet, chen2023teaching, ridnik2024code}, in addition to direct code generation approaches. In contrast, our work combines all these paradigms and bridges their gaps (See Table \ref{tab:feature-compare-table}). Recently, numerous works have explored multi-agent code generation and problem-solving, including \cite{kulesza2004generative, jin2024mare, phan2024hyperagent}, as well as approaches highlighted in Section \ref{sec:intro}. However, \tool uniquely features simulation-driven planning and LLM-based debugging. More recently, external debugging has emerged to further boost performance, such as LDB \cite{zhong-etal-2024-debug}, ChatDebug \cite{levin2024chatdbg}, and MGDebugger \cite{shi2024code-mgdebug}, which serve as a \emph{second pass} after our generation.

\begin{table}[t]
    \centering
    \begin{tabular}{c}
    \hspace*{-0.35cm}
    \includegraphics[width=0.49\textwidth]{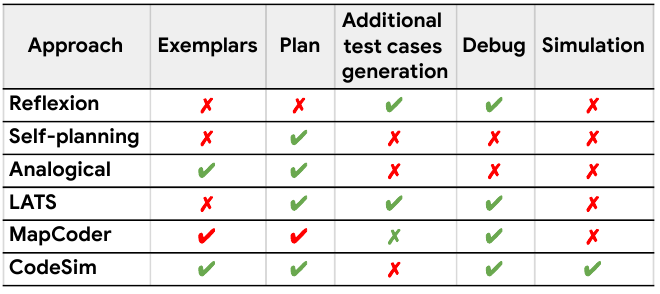}
    \end{tabular}
    \vspace{-3mm}
    \caption{Comparison of code generation approaches.}
    \label{tab:feature-compare-table}
    \vspace{-4mm}
\end{table}

\section{\tool}
\label{sec:mapcoder}
\label{sec:codesim}

Our goal is to develop a multi-agent code generation approach capable of complex problem solving. Drawing inspiration from recent works like MapCoder and ChatDev (in a different context), we devise the agents in \tool for planning, coding, and debugging. 
While these existing approaches focus primarily on expanding steps without verifying underlying hypotheses, we address this limitation by introducing a novel verification approach. Our approach simulates input/output step-by-step, verifying generated plans and performing internal debugging, mirroring how humans understand, visualize, and refine in algorithm development. Below, we present our proposed model.


\begin{table*}[h]
    \centering
    \begin{tabular}{c}
    \hspace{-2.5mm}
    \includegraphics[width=0.999\textwidth]{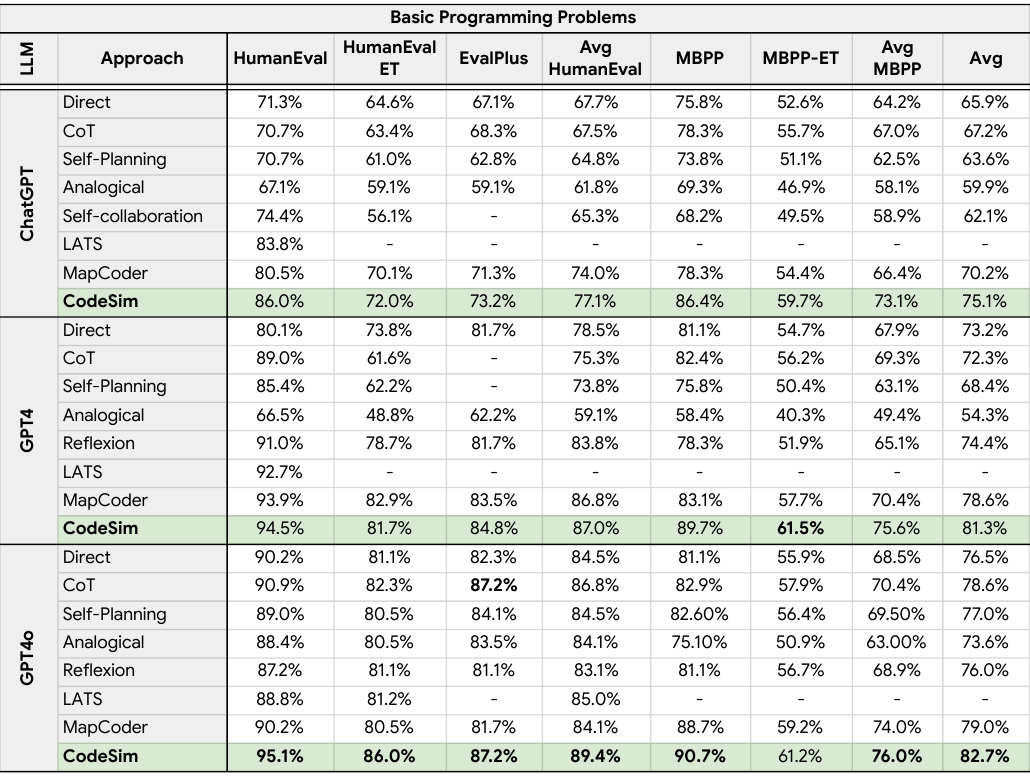}
    \end{tabular}
    \caption{Pass@1 results for different approaches on basic programming tasks.}
    \label{tab:basic-dataset-result}
\end{table*}

\subsection{Planning Agent}
\label{subsec:planning-agent}

The first component of \tool is the \emph{Planning Agent}. Given a problem description, the \emph{Planning Agent} generates a single exemplar—a relevant problem along with its plan and solution. This mimics the behavior of human programmers, who, when faced with a new problem, first recall a similar problem they’ve previously solved. This exemplar-based recall is crucial as it provides a starting point for constructing a solution plan. Instead of generating multiple ungrounded exemplars as in MapCoder, our agent focuses on only one at a time. We then instruct the LLM to 
generate an appropriate plan. Once the plan is created, the LLM simulates (step-by-step) the solution with a sample input. If the simulation result does not match the expected output, the agent prompts the LLM to revise the plan. Otherwise, the plan is deemed valid. In the case of failure, the \emph{Planning Agent} refines the plan. The complete prompts for the Planning Agent—including plan generation, verification, and refinement—are provided in the Appendix (Figure \ref{prompt:plan-generation}, 
\ref{prompt:plan-verification}, \ref{prompt:plan-refinement}).
\subsection{Coding Agent}
\label{subsec:coding-agent}

Next component is the \emph{Coding Agent}, which takes the problem description and the plan generated by the \emph{Planning Agent} as input. The role of this agent is to translate the plan into executable code that solves the given problem. Once the code is generated, \tool evaluates it using sample input/output test cases. If the code passes all sample tests, it is returned as the final solution. Otherwise, the code is handed over to the next agent for further refinement. Figure \ref{prompt:code-generation} in the Appendix provides the complete prompt used by the \emph{Coding Agent}.

\subsection{Debugging Agent}
\label{subsec:debugging-agent}

The final component, the \emph{Debugging Agent}, receives the original problem, the plan from the \emph{Planning Agent}, the code generated by the \emph{Coding Agent}, and the execution (unit testing) log as input to debug the code. To identify bugs, instead of directly prompting the LLMs, we uniquely leverage the simulation once again. The LLM is instructed specifically to simulate the code on inputs where it fails to produce the expected output, allowing it to trace the execution step by step and locate the error. Once the bug is identified, the LLM modifies the code to resolve the issue. The complete prompt for the \emph{Debugging Agent} is shown in the Appendix (Figure \ref{prompt:debugging}). Unlike other approaches such as LATS \cite{zhou2023lats}, AgentCoder \cite{huang2023agentcoder}, and Reflexion \cite{shinn2023reflexion}, our \emph{Debugging Agent} does not require additional test case generation. The rationale behind excluding this phase is discussed in the Ablation Study \ref{subsec:impact-of-additional-io}.



\smallskip
\subsection{Adaptive Iteration}
\label{sec:agent-traverse}

\tool employs an adaptive iteration starting with the \emph{Planning Agent}, which generates plans for the given problem. These plans are passed to the \emph{Coding Agent}, which translates them into code and tests against sample I/Os. If all tests pass, the code is returned; otherwise, it's sent to the \emph{Debugging Agent}. The \emph{Debugging Agent} attempts to fix the code for up to $d$ attempts. If unsuccessful after $d$ attempts, the process returns to the \emph{Planning Agent}, restarting the cycle. Once code passing all sample I/Os is obtained, the cycle ends, returning the code as the final output solution for evaluation against hidden test cases. This entire process repeats for a maximum of $p$ cycles if needed. Algorithm \ref{alg:codesim} in the Appendix summarizes our adaptive agent traversal. The algorithm's complexity is $O(pd)$. Appendix \ref{app:example-problem} provides a comprehensive example of how \tool solves a problem. 

\section{Experimental Setup}
\subsection{Datasets}
Following MapCoder, we evaluate \tool on five basic programming benchmarks i.e., \textbf{HumanEval}~\cite{chen2021codex}, \textbf{HumanEval-ET}~\cite{dong2023codescore}, \textbf{EvalPlus}~\cite{evalplus}, \textbf{MBPP})~\cite{austin2021program}, and \textbf{MBPP-ET}~\cite{dong2023codescore} and two competitive programming datasets i.e., \textbf{APPS} \cite{hendrycks2021apps}, and \textbf{CodeContest} \cite{li2022codecontest}. For fair comparison, we collect all the datasets from the repository of the MapCoder. 

\subsection{Baselines and Metric}
\label{sec:baselines-metrics}
To evaluate \toolnospace, we compare it against state-of-the-art code generation approaches, including {\bf MapCoder}, as well as several notable methods: {\bf Direct}, Chain of Thought  (\textbf{CoT}) \cite{wei2022chain}, \textbf{Self-Planning}  \cite{jiang2023self}, \textbf{Analogical Reasoning}  \cite{yasunaga2023large}, and \textbf{Self-collaboration} \cite{dong2023selfcollaboration}. For simpler programming tasks, we include strong baselines such as Reflexion \cite{shinn2023reflexion} and LATS \cite{zhou2023lats}. We exclude AgentCoder \cite{huang2023agentcoder} due to reproducibility issues (discussed in Appendix \ref{app:elimination-of-agentcoder}). 
For fair comparison, our evaluation utilizes \href{https://platform.openai.com/docs/models/gpt-3-5-turbo}{ChatGPT (gpt-3.5-turbo-1106)}, \href{https://platform.openai.com/docs/models/gpt-4-and-gpt-4-turbo}{GPT-4 (gpt-4-1106-preview)} from OpenAI, alongside open-source LLMs such as Gemma2-9B, Mixtral8x7B, LLaMa3.1-8B, and LLaMa3.1-70B. For basic programming tasks, we report next-generation performance with additional evaluations using  \href{https://platform.openai.com/docs/models/gpt-4o}{GPT-4o (gpt-4o-2024-08-06)}. We adopt the widely used $pass@1$ metric, where a model is deemed successful if its sole predicted solution is correct.

\subsection{Reproducibility}
We aim to contribute to the NLP community by open-sourcing all of our code along with result logs, enabling others to reproduce our findings. For simple programming, we set the maximum number of planning tries to $p = 5$ and debugging tries to $d = 5$. For the competitive problem solving, we used $p = 3$ and $d = 3$ by default except for the CodeContest with GPT-4 where $p = 3$, $d = 5$.

\section{Results}
\label{sec:results}

\subsection{Basic Code Generation}


In Table \ref{tab:basic-dataset-result}, we evaluate the model performances on simple code generation tasks. Overall, \tool demonstrates consistently superior performance compared to all other baselines across all datasets and LLMs. Notably, \tool achieves top scores with GPT-4o, reaching {\bf 95.1\%} on HumanEval, {\bf87.2\%} on EvalPlus, and {\bf90.7\%} on MBPP, resulting in an impressive {\bf82.7\%} overall average and their new state-of-the-art (SoTA)  results. This represents a significant improvement over the next best method, MapCoder, which scores 79.0\% on average with GPT-4o. \toolnospace's effectiveness is consistent across different model variants, outperforming other approaches with ChatGPT ({\bf 75.1\% avg}) and GPT-4 ({\bf 81.3\% avg}) as well. The method's robust performance across diverse datasets, including the challenging MBPP-ET where it achieves {\bf61.5\%} with GPT-4, underscores its versatility in handling various programming tasks. These results strongly indicate that \toolnospace's simulation-driven planning and debugging approach marks a substantial advancement in code generation and problem-solving capabilities, as it consistently outperformed other baselines.

\subsection{ Competitive Problem Solving}

In Table \ref{tab:cc-dataset-results}, we evaluate performance on complex, contest-level code generation tasks. \tool delivers significant improvements over other baselines in solving complex contest-level code generation tasks. With GPT-4, \tool reaches a strong {\bf 29.1\%} on CodeContests and {22.0\%} on APPS, marking a consistent edge over MapCoder’s 25.3\% average. 
The performance gains are even more pronounced with ChatGPT, where \tool achieves a {\bf 16.4\%} on CodeContests, and {12.0\%} on APPS resulting {\bf 14.2\%} overall, outperforming MapCoder’s 12.0\%. These results highlight \toolnospace's ability to handle the complexity of contest-level problems more effectively, especially through its simulation-driven approach.


\begin{table}[h]
    \centering
    \begin{tabular}{c}
    \hspace*{-0.3cm}
    \includegraphics[width=0.48\textwidth]{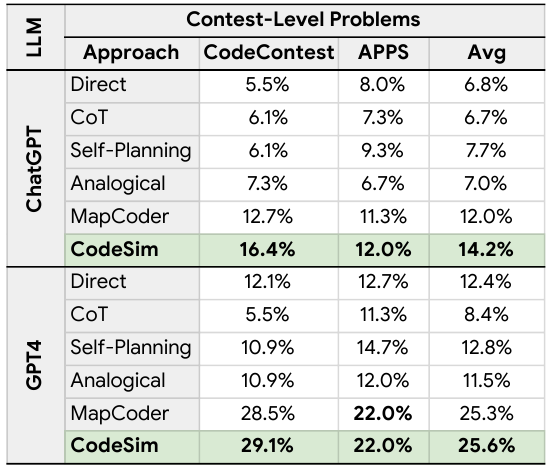}
    \end{tabular}
    \caption{Pass@1 results for different approaches on CodeContest and APPS dataset. }
    \label{tab:cc-dataset-results}
    \vspace{-4mm}
\end{table}

\begin{table}[h]
    \centering
    \begin{tabular}{c}
    \hspace*{-0.35cm}
    \includegraphics[width=0.49\textwidth]{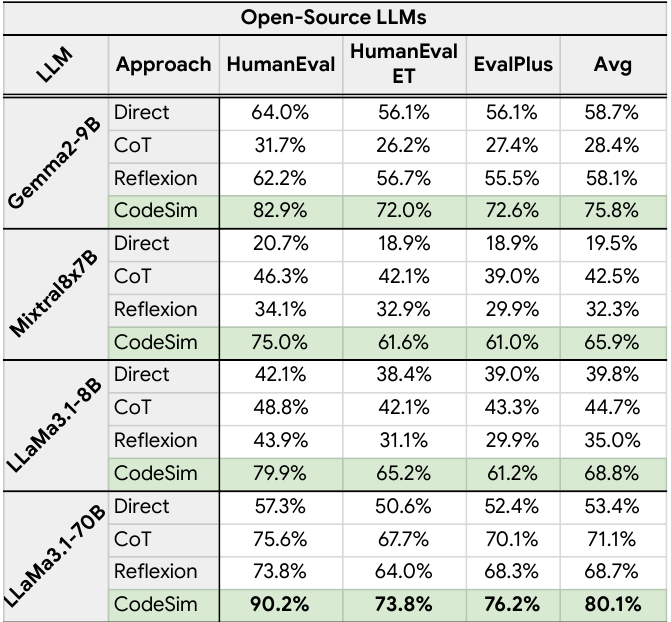}
    \end{tabular}
    \caption{Pass@1 results for different approaches using Open-source LLMs. }
    \label{tab:open-llm-results}
\end{table}

\subsection{Performance Across Open-source LLMs}
\label{sec:exp:open-source}

To further demonstrate \toolnospace’s generalization capability, we evaluate its performance with open-source LLMs, including Gemma2-9B, Mixtral8x7B, LLaMa3.1-8B, and LLaMa3.1-70B. As shown in Table \ref{tab:open-llm-results}, \tool consistently outperforms all other methods across these models. On LLaMa3.1-70B, \tool achieves an accuracy of {\bf 90.2\%} on HumanEval and {\bf 76.2\%} on EvalPlus, with an average of {\bf 80.1\%}, closely matching GPT-4o's performance. Due to the complex prompting scheme of MapCoder, open-source LLMs often struggle to generate output in the correct format. Therefore, we exclude MapCoder from this experiment. On the other hand, Reflexion shows minimal improvement in accuracy. These results highlight \toolnospace's strong generalization ability across various LLM architectures, even on smaller models like Gemma2-9B that achieves a notable avg accuracy of {\bf 75.8\%}.

\section{Ablation Studies and Analyses}
\label{sec:ablation-study}

\subsection{Impact of Different Agents}
\label{sec:agent-impact-ablation}
Our primary contributions are two folds: (i) the simulation-guided plan verification step within the \emph{Planning Agent} and (ii) the bug fixing process through simulation in \emph{Debugging Agent}. To evaluate the significance of these components, we ablate these two parts of our approach and present the results in Table \ref{tab:agent-disabling}. The findings confirm that both components contribute significantly. 

\begin{table}[h]
    \centering
    \begin{tabular}{c}
    \hspace*{-0.35cm}
    \includegraphics[width=0.45\textwidth]{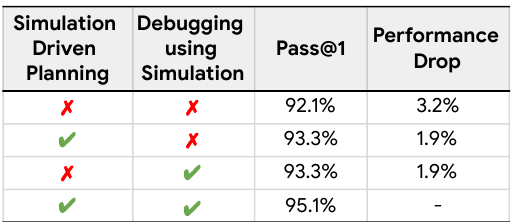}
    \end{tabular}
    \caption{Pass@1 results for different versions of \tool (by using GPT4o on HumanEval dataset).}
    \label{tab:agent-disabling}
\end{table}

\subsection{Fine-grained Analysis of the Impact of Simulation}
\label{subsec:impact-simulation}
Table \ref{tab:simulation-impact} presents the impact of incorporating \textit{Simulation} in \toolnospace. The results show that \tool consistently outperforms other approaches across both simple and multi-agent settings, demonstrating superior performance with both open-source and proprietary LLMs. This highlights the effectiveness of \textit{Simulation} in enhancing problem-solving efficiency within our pipeline. 

\begin{table}[h]
    \centering
    \begin{tabular}{c}
    \hspace*{-0.35cm}
    \includegraphics[width=0.49\textwidth]{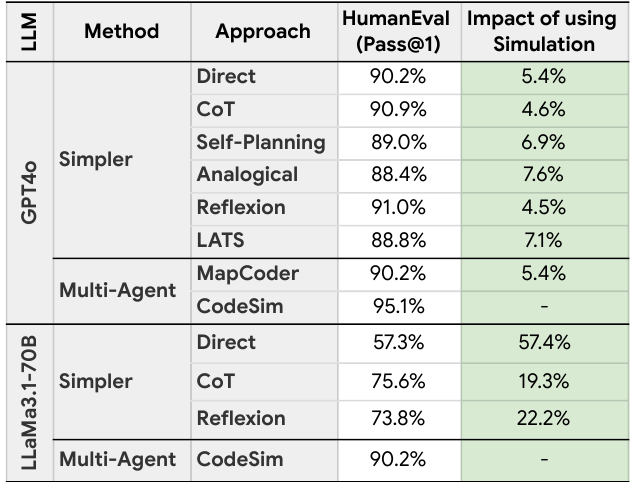}
    \end{tabular}
    \caption{Impact of using \textit{Simulation}.}
    \label{tab:simulation-impact}
\end{table}

\subsection{Impact of Varying Programming Languages}
\label{subsec:various-programming-languages}
To evaluate the performance of \tool across various programming languages, we utilized the xCodeEval \cite{khan2023xcodeeval} dataset. The experimental results, presented in Table \ref{tab:veriying-PL}, demonstrate that \tool maintains strong performance across different programming languages, highlighting its versatility and effectiveness.

\begin{table}[h]
    \centering
    \begin{tabular}{c}    
    \hspace*{-0.35cm}
    \includegraphics[width=0.49\textwidth]{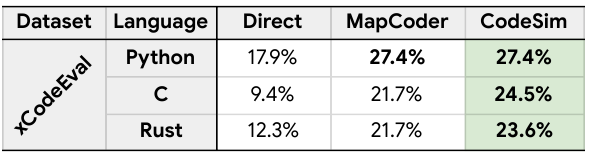}
    \end{tabular}
    \caption{Pass@1 results for different programming languages from xCodeEval dataset by using ChatGPT.}
    \label{tab:veriying-PL}
\end{table}

\subsection{Use of External Debugger}
\label{subsec:ldb-debugger}

\begin{table}[h]
    \centering
    \begin{tabular}{c}
    \hspace*{-0.35cm}
    \includegraphics[width=0.49\textwidth]{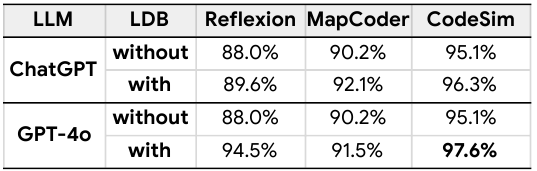}
    \end{tabular}
    \caption{Pass@1 results for different approaches using an external debugger.}
    \label{tab:ldb-results}
\end{table}

\noindent
The performance of \tool can be further enhanced by incorporating an external debugger in the \emph{second pass}. We experiment with LDB as the external debugger on HumanEval dataset in Table \ref{tab:ldb-results}. We use the output code from the most competitive \emph{first-pass} generation methods, including \toolnospace, Reflexion, and MapCoder, using GPT-4o as the backbone. These seed programs are then passed to LDB, which was tested with two different LLMs: ChatGPT and GPT-4o. As can be seen, \tool achieves $95.1$\% accuracy in the \emph{first pass} with GPT-4o, surpassing Reflexion's \emph{second pass} performance of $94.5$\%.  By utilizing LDB with GPT-4o, \tool achieves a \emph{second pass} accuracy of $97.6$\%, setting a new state-of-the-art result for a \emph{dual-pass} approach. In addition, we note that the \emph{second pass} with LDB consumes 39K more tokens in Reflexion compared to our approach, highlighting the efficiency  of \toolnospace.


\subsection{Qualitative Example}
We also conduct a qualitative analysis to better understand how \tool improves performance across various datasets. Figure \ref{fig:qualitative-example} demonstrates how \tool enhances the plan through simulation and assists in debugging the code using the same technique. A complete example, including LLM output, is provided in Appendix \ref{app:example-problem}.

\subsection{Impact of $p$ and $d$}
\begin{figure}[h]
    \centering
    \includegraphics[width=0.85\linewidth]{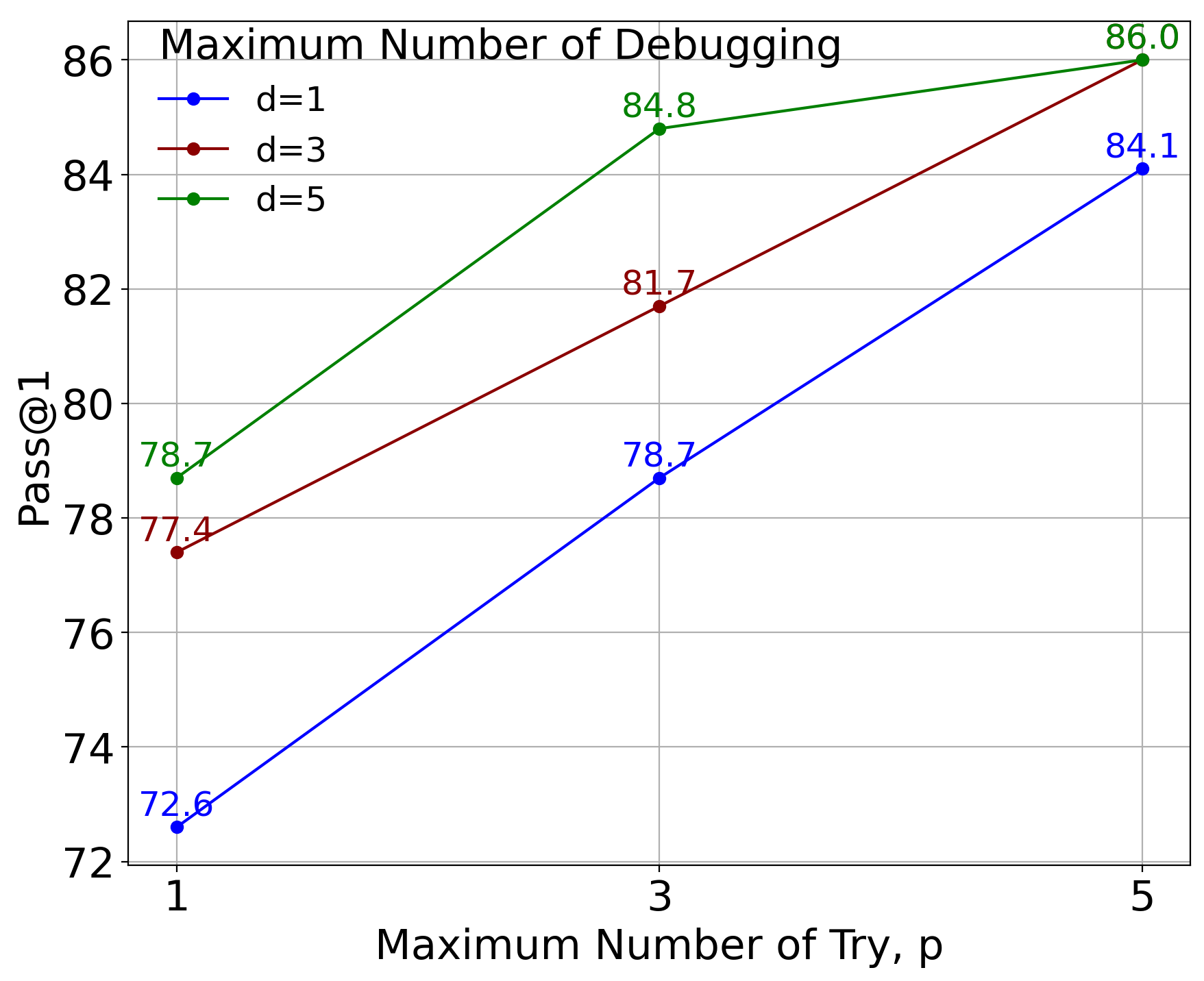}
    \caption{Pass@1 results by varying maximum number of planning, $p$ and maximum number of debugging, $d$.}
    \label{tab:ablation-p-d-results}
\end{figure}

\tool includes two key hyperparameters: the maximum number of planning steps ($p$) and the maximum number of debugging steps ($d$). By varying these parameters, we plot the results in Figure \ref{tab:ablation-p-d-results}, which shows a proportionate improvement in performance. It is important to note that higher values of $p$ and $d$ lead to more API calls and increased token consumption, allowing users to adjust these parameters to balance between accuracy and cost.

\subsection{Impact of Number of Sample I/Os}
\label{subsec:impact-of-sample-io}
The HumanEval dataset has an average of only $2.82$ sample I/Os per example, which is a relatively small number for deriving meaningful insights. In this ablation, we augment the dataset by adding 5 more sample I/Os from the HumanEval-ET dataset. This augmentation increases performance notably, leading to \textbf{$89$\%} accuracy with ChatGPT, a \textbf{$3.5$\%} improvement over previous results, \textbf{$86$\%}.

\begin{table*}[t]
    \centering
    \begin{tabular}{c}
    \hspace*{-0.3cm}
    \includegraphics[width=0.95\textwidth]{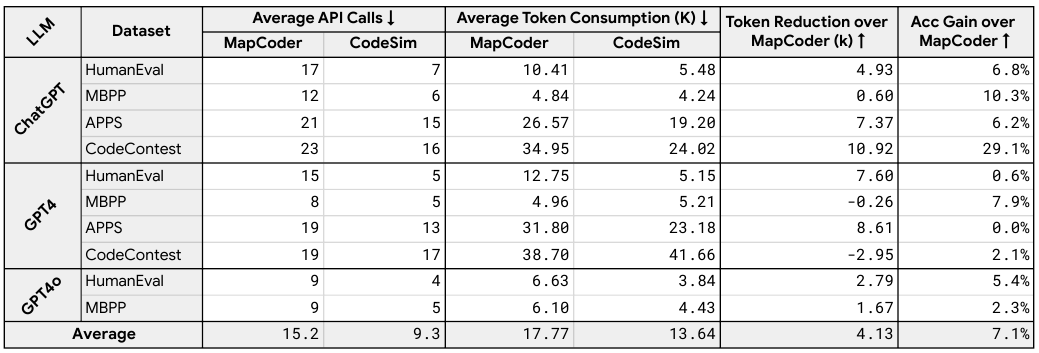}
    \end{tabular}
    \caption{Comparison between MapCoder and \tool in terms of average number of API calls, average tokens used (in thousands). Here the upward symbol (↑) refers that the higher value is better and opposite meaning for downward symbol (↓).}
    \label{tab:ablation-token-counts}
\end{table*}

\subsection{Impact of Synthesizing Additional I/O}
\label{subsec:impact-of-additional-io}
Increasing the number of sample I/Os for testing can enhance the overall performance of our approach, as indicated in \ref{subsec:impact-of-sample-io}. Based on this insight, we use a self-consistency \cite{wang2023selfconsistency} method to generate additional test cases. We instruct the LLM to generate five more test cases for each problem, covering both basic and edge cases. The LLM is called twice, and we select the test cases that are present in both responses. However, this approach results in a performance decline. With ChatGPT we achieve \textbf{$78$\%} accuracy—a \textbf{$9.3$\%} decrease from the original \textbf{$86$\%}. This indicates that generating additional I/Os is a non-trivial task that may negatively impact final outcomes.

\subsection{API Call and Token Analysis}
\label{subsec:api-time-analysis}
We compare the API calls and token consumption of our approach with the previous state-of-the-art method, MapCoder \cite{islam-etal-2024-mapcoder}, as shown in Table \ref{tab:ablation-token-counts}. The results reveal that \tool not only improves performance but also reduces token consumption. On average, \tool uses $4.13$ thousand fewer tokens while achieving a $7.1$\% increase in accuracy, proving that \tool is more efficient in both accuracy and token usage compared to MapCoder.

\subsection{Error Analysis and Challenges}
\begin{figure}[h]
    \centering
    \includegraphics[width=0.49\textwidth]{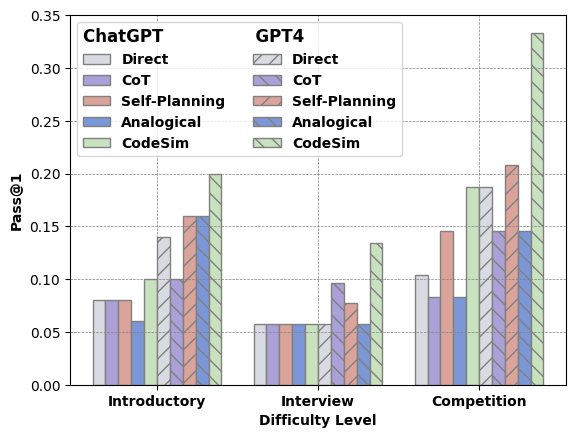}
    \vspace{-3mm}
    \caption{Performance of different approaches across different difficulty levels on the APPS dataset.}
    \label{fig:difficulty-levels}
    \vspace{-4mm}
\end{figure}

\begin{nobreakwords}
Although \tool demonstrates strong performance compared to other methods, it faces challenges in specific algorithmic domains. The APPS dataset \cite{hendrycks2021apps} includes problems with three levels of difficulty: (i) Introductory, (ii) Interview, and (iii) Competition. Figure \ref{fig:difficulty-levels} illustrates the performance of different approaches based on difficulty level. The results indicate that for introductory and interview-level problems, \tool does not surpass MapCoder when using ChatGPT. Additionally, when using GPT-4, \tool struggles to outperform MapCoder on interview-level problems. Upon manual review, we observe that for more complex issues, such as dynamic programming (DP), \tool encounters difficulties in constructing the DP table.
\end{nobreakwords}

\section{Conclusion and Future Work}
\begin{nobreakwords}
In this paper, we introduce \toolnospace, a novel framework that leverages the multi-agent prompting capabilities of LLMs for efficient code generation in problem-solving tasks. \tool integrates three agents—planning, coding, and debugging—to effectively solve programming problems. It harnesses the power of simulation for plan verification and debugging, significantly outperforming existing state-of-the-art approaches by a wide margin. Future work will focus on extending this approach to other domains such as mathematical reasoning and question answering broadening its scope and impact.
\end{nobreakwords}


\section{Limitations}
In Section \ref{subsec:ldb-debugger}, we observe that utilizing an external debugger can further enhance our results. Our next research goal is to achieve the best performance without relying on any external tools. Although we have reduced token consumption compared to the previous state-of-the-art method, MapCoder, it still remains high compared to the direct prompting approach. Direct prompting consumes an average of 560 tokens, while our method consumes around 13,640 tokens. This indicates room for enhancement in efficiency. While in this work, we generate the exemplars with the LLMs themselves, in general they are found from external resource \cite{parvez-chang-2021-evaluating}. Although this has its own challenges such as noisy retrievals \cite{filco}, inconsistent generations \cite{islam-etal-2024-open, parvez2024evidence, sadat-etal-2023-delucionqa} this direction could also be a possible improvement. Another limitation is the use of external tools for assistance during simulation. We have not explored this avenue in the current research, leaving it for future work. Additionally, more sample I/Os could potentially improve performance, and our future research will focus on investigating methods for generating accurate additional I/Os. Moreover, we would like to note that in this work, we focus solely on generated code's correctness and did not study its optimizations such as test-time, memory. Finally, it is advisable to run the machine generated code inside a sandbox to avoid any potential risks.


\bibliography{custom, anthology, iclr2025_conference}

\begin{thebibliography}{70}
\expandafter\ifx\csname natexlab\endcsname\relax\def\natexlab#1{#1}\fi

\bibitem[{Ahmad et~al.(2021)Ahmad, Chakraborty, Ray, and Chang}]{ahmad2021unified}
Wasi~Uddin Ahmad, Saikat Chakraborty, Baishakhi Ray, and Kai-Wei Chang. 2021.
\newblock Unified pre-training for program understanding and generation.
\newblock \emph{arXiv preprint arXiv:2103.06333}.

\bibitem[{Allal et~al.(2023)Allal, Li, Kocetkov, Mou, Akiki, Ferrandis, Muennighoff, Mishra, Gu, Dey et~al.}]{allal2023santacoder}
Loubna~Ben Allal, Raymond Li, Denis Kocetkov, Chenghao Mou, Christopher Akiki, Carlos~Munoz Ferrandis, Niklas Muennighoff, Mayank Mishra, Alex Gu, Manan Dey, et~al. 2023.
\newblock Santacoder: don't reach for the stars!
\newblock \emph{arXiv preprint arXiv:2301.03988}.

\bibitem[{Andreas et~al.(2020)Andreas, Bufe, Burkett, Chen, Clausman, Crawford, Crim, DeLoach, Dorner, Eisner, Fang, Guo, Hall, Hayes, Hill, Ho, Iwaszuk, Jha, Klein, Krishnamurthy, Lanman, Liang, Lin, Lintsbakh, McGovern, Nisnevich, Pauls, Petters, Read, Roth, Roy, Rusak, Short, Slomin, Snyder, Striplin, Su, Tellman, Thomson, Vorobev, Witoszko, Wolfe, Wray, Zhang, and Zotov}]{andreas-etal-2020-task}
Jacob Andreas, John Bufe, David Burkett, Charles Chen, Josh Clausman, Jean Crawford, Kate Crim, Jordan DeLoach, Leah Dorner, Jason Eisner, Hao Fang, Alan Guo, David Hall, Kristin Hayes, Kellie Hill, Diana Ho, Wendy Iwaszuk, Smriti Jha, Dan Klein, Jayant Krishnamurthy, Theo Lanman, Percy Liang, Christopher~H. Lin, Ilya Lintsbakh, Andy McGovern, Aleksandr Nisnevich, Adam Pauls, Dmitrij Petters, Brent Read, Dan Roth, Subhro Roy, Jesse Rusak, Beth Short, Div Slomin, Ben Snyder, Stephon Striplin, Yu~Su, Zachary Tellman, Sam Thomson, Andrei Vorobev, Izabela Witoszko, Jason Wolfe, Abby Wray, Yuchen Zhang, and Alexander Zotov. 2020.
\newblock \href {https://doi.org/10.1162/tacl_a_00333} {Task-oriented dialogue as dataflow synthesis}.
\newblock \emph{Transactions of the Association for Computational Linguistics}, 8:556--571.

\bibitem[{Austin et~al.(2021)Austin, Odena, Nye, Bosma, Michalewski, Dohan, Jiang, Cai, Terry, Le et~al.}]{austin2021program}
Jacob Austin, Augustus Odena, Maxwell Nye, Maarten Bosma, Henryk Michalewski, David Dohan, Ellen Jiang, Carrie Cai, Michael Terry, Quoc Le, et~al. 2021.
\newblock Program synthesis with large language models.
\newblock \emph{arXiv preprint arXiv:2108.07732}.

\bibitem[{Chen et~al.(2022{\natexlab{a}})Chen, Zhang, Nguyen, Zan, Lin, Lou, and Chen}]{codet}
Bei Chen, Fengji Zhang, Anh Nguyen, Daoguang Zan, Zeqi Lin, Jian-Guang Lou, and Weizhu Chen. 2022{\natexlab{a}}.
\newblock Codet: Code generation with generated tests.
\newblock \emph{arXiv preprint arXiv:2207.10397}.

\bibitem[{Chen et~al.(2022{\natexlab{b}})Chen, Zhang, Nguyen, Zan, Lin, Lou, and Chen}]{chen2022codet}
Bei Chen, Fengji Zhang, Anh Nguyen, Daoguang Zan, Zeqi Lin, Jian-Guang Lou, and Weizhu Chen. 2022{\natexlab{b}}.
\newblock Codet: Code generation with generated tests.
\newblock \emph{arXiv preprint arXiv:2207.10397}.

\bibitem[{Chen et~al.(2021{\natexlab{a}})Chen, Tworek, Jun, Yuan, de~Oliveira~Pinto, Kaplan, Edwards, Burda, Joseph, Brockman, Ray, Puri, Krueger, Petrov, Khlaaf, Sastry, Mishkin, Chan, Gray, Ryder, Pavlov, Power, Kaiser, Bavarian, Winter, Tillet, Such, Cummings, Plappert, Chantzis, Barnes, Herbert-Voss, Guss, Nichol, Paino, Tezak, Tang, Babuschkin, Balaji, Jain, Saunders, Hesse, Carr, Leike, Achiam, Misra, Morikawa, Radford, Knight, Brundage, Murati, Mayer, Welinder, McGrew, Amodei, McCandlish, Sutskever, and Zaremba}]{chen2021codex}
Mark Chen, Jerry Tworek, Heewoo Jun, Qiming Yuan, Henrique~Ponde de~Oliveira~Pinto, Jared Kaplan, Harri Edwards, Yuri Burda, Nicholas Joseph, Greg Brockman, Alex Ray, Raul Puri, Gretchen Krueger, Michael Petrov, Heidy Khlaaf, Girish Sastry, Pamela Mishkin, Brooke Chan, Scott Gray, Nick Ryder, Mikhail Pavlov, Alethea Power, Lukasz Kaiser, Mohammad Bavarian, Clemens Winter, Philippe Tillet, Felipe~Petroski Such, Dave Cummings, Matthias Plappert, Fotios Chantzis, Elizabeth Barnes, Ariel Herbert-Voss, William~Hebgen Guss, Alex Nichol, Alex Paino, Nikolas Tezak, Jie Tang, Igor Babuschkin, Suchir Balaji, Shantanu Jain, William Saunders, Christopher Hesse, Andrew~N. Carr, Jan Leike, Josh Achiam, Vedant Misra, Evan Morikawa, Alec Radford, Matthew Knight, Miles Brundage, Mira Murati, Katie Mayer, Peter Welinder, Bob McGrew, Dario Amodei, Sam McCandlish, Ilya Sutskever, and Wojciech Zaremba. 2021{\natexlab{a}}.
\newblock \href {http://arxiv.org/abs/2107.03374} {Evaluating large language models trained on code}.

\bibitem[{Chen et~al.(2021{\natexlab{b}})Chen, Tworek, Jun, Yuan, Pinto, Kaplan, Edwards, Burda, Joseph, Brockman et~al.}]{chen2021evaluating}
Mark Chen, Jerry Tworek, Heewoo Jun, Qiming Yuan, Henrique Ponde de~Oliveira Pinto, Jared Kaplan, Harri Edwards, Yuri Burda, Nicholas Joseph, Greg Brockman, et~al. 2021{\natexlab{b}}.
\newblock Evaluating large language models trained on code.
\newblock \emph{arXiv preprint arXiv:2107.03374}.

\bibitem[{Chen et~al.(2023)Chen, Lin, Sch{\"a}rli, and Zhou}]{chen2023teaching}
Xinyun Chen, Maxwell Lin, Nathanael Sch{\"a}rli, and Denny Zhou. 2023.
\newblock Teaching large language models to self-debug.
\newblock \emph{arXiv preprint arXiv:2304.05128}.

\bibitem[{Dong et~al.(2023{\natexlab{a}})Dong, Ding, Jiang, Li, Li, and Jin}]{dong2023codescore}
Yihong Dong, Jiazheng Ding, Xue Jiang, Zhuo Li, Ge~Li, and Zhi Jin. 2023{\natexlab{a}}.
\newblock Codescore: Evaluating code generation by learning code execution.
\newblock \emph{arXiv preprint arXiv:2301.09043}.

\bibitem[{Dong et~al.(2023{\natexlab{b}})Dong, Jiang, Jin, and Li}]{dong2023selfcollaboration}
Yihong Dong, Xue Jiang, Zhi Jin, and Ge~Li. 2023{\natexlab{b}}.
\newblock \href {http://arxiv.org/abs/2304.07590} {Self-collaboration code generation via chatgpt}.

\bibitem[{Dubey et~al.(2024)Dubey, Jauhri, Pandey, Kadian, Al-Dahle, Letman, Mathur, Schelten, Yang, Fan, and et~al.}]{dubey2024llama3herdmodels}
Abhimanyu Dubey, Abhinav Jauhri, Abhinav Pandey, Abhishek Kadian, Ahmad Al-Dahle, Aiesha Letman, Akhil Mathur, Alan Schelten, Amy Yang, Angela Fan, and et~al. 2024.
\newblock \href {http://arxiv.org/abs/2407.21783} {The llama 3 herd of models}.

\bibitem[{Feng et~al.(2020)Feng, Guo, Tang, Duan, Feng, Gong, Shou, Qin, Liu, Jiang et~al.}]{feng2020codebert}
Zhangyin Feng, Daya Guo, Duyu Tang, Nan Duan, Xiaocheng Feng, Ming Gong, Linjun Shou, Bing Qin, Ting Liu, Daxin Jiang, et~al. 2020.
\newblock Codebert: A pre-trained model for programming and natural languages.
\newblock In \emph{Findings of the Association for Computational Linguistics: EMNLP 2020}, pages 1536--1547.

\bibitem[{Fried et~al.(2022)Fried, Aghajanyan, Lin, Wang, Wallace, Shi, Zhong, Yih, Zettlemoyer, and Lewis}]{fried2022incoder}
Daniel Fried, Armen Aghajanyan, Jessy Lin, Sida Wang, Eric Wallace, Freda Shi, Ruiqi Zhong, Wen-tau Yih, Luke Zettlemoyer, and Mike Lewis. 2022.
\newblock Incoder: A generative model for code infilling and synthesis.
\newblock \emph{arXiv preprint arXiv:2204.05999}.

\bibitem[{Gulwani(2011)}]{gulwani2011automating}
Sumit Gulwani. 2011.
\newblock Automating string processing in spreadsheets using input-output examples.
\newblock \emph{ACM Sigplan Notices}, 46(1):317--330.

\bibitem[{Guo et~al.(2024)Guo, Zhu, Yang, Xie, Dong, Zhang, Chen, Bi, Wu, Li et~al.}]{guo2024deepseek}
Daya Guo, Qihao Zhu, Dejian Yang, Zhenda Xie, Kai Dong, Wentao Zhang, Guanting Chen, Xiao Bi, Y~Wu, YK~Li, et~al. 2024.
\newblock Deepseek-coder: When the large language model meets programming--the rise of code intelligence.
\newblock \emph{arXiv preprint arXiv:2401.14196}.

\bibitem[{Hellendoorn and Devanbu(2017)}]{deep_net_for_source_code}
Vincent~J. Hellendoorn and Premkumar Devanbu. 2017.
\newblock \href {https://doi.org/10.1145/3106237.3106290} {Are deep neural networks the best choice for modeling source code?}
\newblock In \emph{Proceedings of the 2017 11th Joint Meeting on Foundations of Software Engineering}, ESEC/FSE 2017, pages 763--773, New York, NY, USA. ACM.

\bibitem[{Hendrycks et~al.(2021)Hendrycks, Basart, Kadavath, Mazeika, Arora, Guo, Burns, Puranik, He, Song et~al.}]{hendrycks2021apps}
Dan Hendrycks, Steven Basart, Saurav Kadavath, Mantas Mazeika, Akul Arora, Ethan Guo, Collin Burns, Samir Puranik, Horace He, Dawn Song, et~al. 2021.
\newblock Measuring coding challenge competence with apps.
\newblock \emph{arXiv preprint arXiv:2105.09938}.

\bibitem[{Hindle et~al.(2016)Hindle, Barr, Gabel, Su, and Devanbu}]{naturalnessofsoft}
Abram Hindle, Earl~T. Barr, Mark Gabel, Zhendong Su, and Premkumar Devanbu. 2016.
\newblock \href {https://doi.org/10.1145/2902362} {On the naturalness of software}.
\newblock \emph{Commun. ACM}, 59(5):122--131.

\bibitem[{Huang et~al.(2023)Huang, Bu, Zhang, Luck, and Cui}]{huang2023agentcoder}
Dong Huang, Qingwen Bu, Jie~M Zhang, Michael Luck, and Heming Cui. 2023.
\newblock Agentcoder: Multi-agent-based code generation with iterative testing and optimisation.
\newblock \emph{arXiv preprint arXiv:2312.13010}.

\bibitem[{Hui et~al.(2024)Hui, Yang, Cui, Yang, Liu, Zhang, Liu, Zhang, Yu, Dang et~al.}]{hui2024qwen2}
Binyuan Hui, Jian Yang, Zeyu Cui, Jiaxi Yang, Dayiheng Liu, Lei Zhang, Tianyu Liu, Jiajun Zhang, Bowen Yu, Kai Dang, et~al. 2024.
\newblock Qwen2. 5-coder technical report.
\newblock \emph{arXiv preprint arXiv:2409.12186}.

\bibitem[{Islam et~al.(2024{\natexlab{a}})Islam, Ali, and Parvez}]{islam-etal-2024-mapcoder}
Md.~Ashraful Islam, Mohammed~Eunus Ali, and Md~Rizwan Parvez. 2024{\natexlab{a}}.
\newblock \href {https://doi.org/10.18653/v1/2024.acl-long.269} {{M}ap{C}oder: Multi-agent code generation for competitive problem solving}.
\newblock In \emph{Proceedings of the 62nd Annual Meeting of the Association for Computational Linguistics (Volume 1: Long Papers)}, pages 4912--4944, Bangkok, Thailand. Association for Computational Linguistics.

\bibitem[{Islam et~al.(2024{\natexlab{b}})Islam, Rahman, Hossain, Hoque, Joty, and Parvez}]{islam-etal-2024-open}
Shayekh~Bin Islam, Md~Asib Rahman, K~S M~Tozammel Hossain, Enamul Hoque, Shafiq Joty, and Md~Rizwan Parvez. 2024{\natexlab{b}}.
\newblock \href {https://doi.org/10.18653/v1/2024.findings-emnlp.831} {Open-{RAG}: Enhanced retrieval augmented reasoning with open-source large language models}.
\newblock In \emph{Findings of the Association for Computational Linguistics: EMNLP 2024}, pages 14231--14244, Miami, Florida, USA. Association for Computational Linguistics.

\bibitem[{Jiang et~al.(2023{\natexlab{a}})Jiang, Sablayrolles, Mensch, Bamford, Chaplot, de~las Casas, Bressand, Lengyel, Lample, Saulnier, Lavaud, Lachaux, Stock, Scao, Lavril, Wang, Lacroix, and Sayed}]{jiang2023mistral}
Albert~Q. Jiang, Alexandre Sablayrolles, Arthur Mensch, Chris Bamford, Devendra~Singh Chaplot, Diego de~las Casas, Florian Bressand, Gianna Lengyel, Guillaume Lample, Lucile Saulnier, Lélio~Renard Lavaud, Marie-Anne Lachaux, Pierre Stock, Teven~Le Scao, Thibaut Lavril, Thomas Wang, Timothée Lacroix, and William~El Sayed. 2023{\natexlab{a}}.
\newblock \href {http://arxiv.org/abs/2310.06825} {Mistral 7b}.

\bibitem[{Jiang et~al.(2023{\natexlab{b}})Jiang, Dong, Wang, Shang, and Li}]{jiang2023self}
Xue Jiang, Yihong Dong, Lecheng Wang, Qiwei Shang, and Ge~Li. 2023{\natexlab{b}}.
\newblock Self-planning code generation with large language model.
\newblock \emph{arXiv preprint arXiv:2303.06689}.

\bibitem[{Jin et~al.(2024{\natexlab{a}})Jin, Jin, Chen, and Wang}]{jin2024mare}
Dongming Jin, Zhi Jin, Xiaohong Chen, and Chunhui Wang. 2024{\natexlab{a}}.
\newblock Mare: Multi-agents collaboration framework for requirements engineering.
\newblock \emph{arXiv preprint arXiv:2405.03256}.

\bibitem[{Jin et~al.(2024{\natexlab{b}})Jin, Sun, Yang, and Chen}]{jin2024rgd}
Haolin Jin, Zechao Sun, Yiheng Yang, and Huaming Chen. 2024{\natexlab{b}}.
\newblock Rgd: Multi-llm based agent debugger via refinement and generation guidance.
\newblock \emph{arXiv preprint arXiv:2410.01242}.

\bibitem[{Khan et~al.(2023)Khan, Bari, Do, Wang, Parvez, and Joty}]{khan2023xcodeeval}
Mohammad Abdullah~Matin Khan, M~Saiful Bari, Xuan~Long Do, Weishi Wang, Md~Rizwan Parvez, and Shafiq Joty. 2023.
\newblock xcodeeval: A large scale multilingual multitask benchmark for code understanding, generation, translation and retrieval.
\newblock \emph{arXiv preprint arXiv:2303.03004}.

\bibitem[{Kulesza et~al.(2004)Kulesza, Garcia, Lucena, and Alencar}]{kulesza2004generative}
Uir{\'a} Kulesza, Alessandro Garcia, Carlos Lucena, and Paulo Alencar. 2004.
\newblock A generative approach for multi-agent system development.
\newblock In \emph{International Workshop on Software Engineering for Large-Scale Multi-agent Systems}, pages 52--69. Springer.

\bibitem[{Laskar et~al.(2024)Laskar, Alqahtani, Bari, Rahman, Khan, Khan, Jahan, Bhuiyan, Tan, Parvez, Hoque, Joty, and Huang}]{laskar-etal-2024-systematic}
Md~Tahmid~Rahman Laskar, Sawsan Alqahtani, M~Saiful Bari, Mizanur Rahman, Mohammad Abdullah~Matin Khan, Haidar Khan, Israt Jahan, Amran Bhuiyan, Chee~Wei Tan, Md~Rizwan Parvez, Enamul Hoque, Shafiq Joty, and Jimmy Huang. 2024.
\newblock \href {https://doi.org/10.18653/v1/2024.emnlp-main.764} {A systematic survey and critical review on evaluating large language models: Challenges, limitations, and recommendations}.
\newblock In \emph{Proceedings of the 2024 Conference on Empirical Methods in Natural Language Processing}, pages 13785--13816, Miami, Florida, USA. Association for Computational Linguistics.

\bibitem[{Le et~al.(2022)Le, Wang, Gotmare, Savarese, and Hoi}]{le2022coderl}
Hung Le, Yue Wang, Akhilesh~Deepak Gotmare, Silvio Savarese, and Steven Chu~Hong Hoi. 2022.
\newblock Coderl: Mastering code generation through pretrained models and deep reinforcement learning.
\newblock \emph{Advances in Neural Information Processing Systems}, 35:21314--21328.

\bibitem[{Levin et~al.(2024)Levin, van Kempen, Berger, and Freund}]{levin2024chatdbg}
Kyla Levin, Nicolas van Kempen, Emery~D Berger, and Stephen~N Freund. 2024.
\newblock Chatdbg: An ai-powered debugging assistant.
\newblock \emph{arXiv preprint arXiv:2403.16354}.

\bibitem[{Li et~al.(2023)Li, Chen, and Jia}]{li2023motcoder}
Jingyao Li, Pengguang Chen, and Jiaya Jia. 2023.
\newblock Motcoder: Elevating large language models with modular of thought for challenging programming tasks.
\newblock \emph{arXiv preprint arXiv:2312.15960}.

\bibitem[{Li et~al.(2022{\natexlab{a}})Li, Choi, Chung, Kushman, Schrittwieser, Leblond, Eccles, Keeling, Gimeno, Dal~Lago et~al.}]{li2022competition}
Yujia Li, David Choi, Junyoung Chung, Nate Kushman, Julian Schrittwieser, R{\'e}mi Leblond, Tom Eccles, James Keeling, Felix Gimeno, Agustin Dal~Lago, et~al. 2022{\natexlab{a}}.
\newblock Competition-level code generation with alphacode.
\newblock \emph{Science}, 378(6624):1092--1097.

\bibitem[{Li et~al.(2022{\natexlab{b}})Li, Choi, Chung, Kushman, Schrittwieser, Leblond, Eccles, Keeling, Gimeno, Lago, Hubert, Choy, de~Masson~d’Autume, Babuschkin, Chen, Huang, Welbl, Gowal, Cherepanov, Molloy, Mankowitz, Robson, Kohli, de~Freitas, Kavukcuoglu, and Vinyals}]{li2022codecontest}
Yujia Li, David Choi, Junyoung Chung, Nate Kushman, Julian Schrittwieser, R{\'e}mi Leblond, Tom Eccles, James Keeling, Felix Gimeno, Agustin~Dal Lago, Thomas Hubert, Peter Choy, Cyprien de~Masson~d’Autume, Igor Babuschkin, Xinyun Chen, Po-Sen Huang, Johannes Welbl, Sven Gowal, Alexey Cherepanov, James Molloy, Daniel~J. Mankowitz, Esme~Sutherland Robson, Pushmeet Kohli, Nando de~Freitas, Koray Kavukcuoglu, and Oriol Vinyals. 2022{\natexlab{b}}.
\newblock \href {https://doi.org/10.1126/science.abq1158} {Competition-level code generation with alphacode}.
\newblock \emph{Science}, 378(6624):1092--1097.

\bibitem[{Li et~al.(2022{\natexlab{c}})Li, Choi, Chung, Kushman, Schrittwieser, Leblond, Eccles, Keeling, Gimeno, Dal~Lago, Hubert, Choy, de~Masson~d'Autume, Babuschkin, Chen, Huang, Welbl, Gowal, Cherepanov, Molloy, Mankowitz, Sutherland~Robson, Kohli, de~Freitas, Kavukcuoglu, and Vinyals}]{alphacode}
Yujia Li, David Choi, Junyoung Chung, Nate Kushman, Julian Schrittwieser, Rémi Leblond, Tom Eccles, James Keeling, Felix Gimeno, Agustin Dal~Lago, Thomas Hubert, Peter Choy, Cyprien de~Masson~d'Autume, Igor Babuschkin, Xinyun Chen, Po-Sen Huang, Johannes Welbl, Sven Gowal, Alexey Cherepanov, James Molloy, Daniel Mankowitz, Esme Sutherland~Robson, Pushmeet Kohli, Nando de~Freitas, Koray Kavukcuoglu, and Oriol Vinyals. 2022{\natexlab{c}}.
\newblock Competition-level code generation with alphacode.

\bibitem[{Liu et~al.(2023)Liu, Xia, Wang, and Zhang}]{evalplus}
Jiawei Liu, Chunqiu~Steven Xia, Yuyao Wang, and Lingming Zhang. 2023.
\newblock \href {https://openreview.net/forum?id=1qvx610Cu7} {Is your code generated by chat{GPT} really correct? rigorous evaluation of large language models for code generation}.
\newblock In \emph{Thirty-seventh Conference on Neural Information Processing Systems}.

\bibitem[{Manna and Waldinger(1971)}]{Zohar71}
Zohar Manna and Richard~J. Waldinger. 1971.
\newblock \href {https://doi.org/10.1145/362566.362568} {Toward automatic program synthesis}.
\newblock \emph{Commun. ACM}, 14(3):151–165.

\bibitem[{Nijkamp et~al.(2022)Nijkamp, Pang, Hayashi, Tu, Wang, Zhou, Savarese, and Xiong}]{nijkamp2022codegen}
Erik Nijkamp, Bo~Pang, Hiroaki Hayashi, Lifu Tu, Huan Wang, Yingbo Zhou, Silvio Savarese, and Caiming Xiong. 2022.
\newblock Codegen: An open large language model for code with multi-turn program synthesis.
\newblock \emph{arXiv preprint arXiv:2203.13474}.

\bibitem[{OpenAI(2024)}]{openai2024gpt4technicalreport}
OpenAI. 2024.
\newblock \href {http://arxiv.org/abs/2303.08774} {Gpt-4 technical report}.

\bibitem[{Parisotto and Salakhutdinov(2017)}]{parisotto2017neural}
Emilio Parisotto and Ruslan Salakhutdinov. 2017.
\newblock Neural map: Structured memory for deep reinforcement learning.
\newblock \emph{arXiv preprint arXiv:1702.08360}.

\bibitem[{Parvez(2024)}]{parvez2024evidence}
Md~Rizwan Parvez. 2024.
\newblock Evidence to generate (e2g): A single-agent two-step prompting for context grounded and retrieval augmented reasoning.
\newblock \emph{arXiv preprint arXiv:2401.05787}.

\bibitem[{Parvez et~al.(2021)Parvez, Ahmad, Chakraborty, Ray, and Chang}]{parvez2021retrieval}
Md~Rizwan Parvez, Wasi~Uddin Ahmad, Saikat Chakraborty, Baishakhi Ray, and Kai-Wei Chang. 2021.
\newblock Retrieval augmented code generation and summarization.
\newblock \emph{arXiv preprint arXiv:2108.11601}.

\bibitem[{Parvez et~al.(2018)Parvez, Chakraborty, Ray, and Chang}]{parvez-etal-2018-building}
Md~Rizwan Parvez, Saikat Chakraborty, Baishakhi Ray, and Kai-Wei Chang. 2018.
\newblock \href {https://doi.org/10.18653/v1/P18-1221} {Building language models for text with named entities}.
\newblock In \emph{Proceedings of the 56th Annual Meeting of the Association for Computational Linguistics (Volume 1: Long Papers)}, pages 2373--2383, Melbourne, Australia. Association for Computational Linguistics.

\bibitem[{Parvez and Chang(2021)}]{parvez-chang-2021-evaluating}
Md~Rizwan Parvez and Kai-Wei Chang. 2021.
\newblock \href {https://doi.org/10.18653/v1/2021.naacl-main.402} {Evaluating the values of sources in transfer learning}.
\newblock In \emph{Proceedings of the 2021 Conference of the North American Chapter of the Association for Computational Linguistics: Human Language Technologies}, pages 5084--5116, Online. Association for Computational Linguistics.

\bibitem[{Parvez et~al.(2023)Parvez, Chi, Ahmad, Tian, and Chang}]{parvez-etal-2023-retrieval}
Md~Rizwan Parvez, Jianfeng Chi, Wasi~Uddin Ahmad, Yuan Tian, and Kai-Wei Chang. 2023.
\newblock \href {https://doi.org/10.18653/v1/2023.eacl-main.16} {Retrieval enhanced data augmentation for question answering on privacy policies}.
\newblock In \emph{Proceedings of the 17th Conference of the European Chapter of the Association for Computational Linguistics}, pages 201--210, Dubrovnik, Croatia. Association for Computational Linguistics.

\bibitem[{Phan et~al.(2024)Phan, Nguyen, and Bui}]{phan2024hyperagent}
Huy~Nhat Phan, Phong~X Nguyen, and Nghi~DQ Bui. 2024.
\newblock Hyperagent: Generalist software engineering agents to solve coding tasks at scale.
\newblock \emph{arXiv preprint arXiv:2409.16299}.

\bibitem[{Polozov and Gulwani(2015)}]{polozov2015flashmeta}
Oleksandr Polozov and Sumit Gulwani. 2015.
\newblock Flashmeta: A framework for inductive program synthesis.
\newblock In \emph{Proceedings of the 2015 ACM SIGPLAN International Conference on Object-Oriented Programming, Systems, Languages, and Applications}, pages 107--126.

\bibitem[{Qian et~al.(2024)Qian, Liu, Liu, Chen, Dang, Li, Yang, Chen, Su, Cong, Xu, Li, Liu, and Sun}]{qian-etal-2024-chatdev}
Chen Qian, Wei Liu, Hongzhang Liu, Nuo Chen, Yufan Dang, Jiahao Li, Cheng Yang, Weize Chen, Yusheng Su, Xin Cong, Juyuan Xu, Dahai Li, Zhiyuan Liu, and Maosong Sun. 2024.
\newblock \href {https://doi.org/10.18653/v1/2024.acl-long.810} {{C}hat{D}ev: Communicative agents for software development}.
\newblock In \emph{Proceedings of the 62nd Annual Meeting of the Association for Computational Linguistics (Volume 1: Long Papers)}, pages 15174--15186, Bangkok, Thailand. Association for Computational Linguistics.

\bibitem[{Rabinovich et~al.(2017)Rabinovich, Stern, and Klein}]{code_gen_parsing}
Maxim Rabinovich, Mitchell Stern, and Dan Klein. 2017.
\newblock \href {http://arxiv.org/abs/1704.07535} {Abstract syntax networks for code generation and semantic parsing}.
\newblock \emph{CoRR}, abs/1704.07535.

\bibitem[{Ridnik et~al.(2024)Ridnik, Kredo, and Friedman}]{ridnik2024code}
Tal Ridnik, Dedy Kredo, and Itamar Friedman. 2024.
\newblock Code generation with alphacodium: From prompt engineering to flow engineering.
\newblock \emph{arXiv preprint arXiv:2401.08500}.

\bibitem[{Roziere et~al.(2023)Roziere, Gehring, Gloeckle, Sootla, Gat, Tan, Adi, Liu, Remez, Rapin et~al.}]{roziere2023code}
Baptiste Roziere, Jonas Gehring, Fabian Gloeckle, Sten Sootla, Itai Gat, Xiaoqing~Ellen Tan, Yossi Adi, Jingyu Liu, Tal Remez, J{\'e}r{\'e}my Rapin, et~al. 2023.
\newblock Code llama: Open foundation models for code.
\newblock \emph{arXiv preprint arXiv:2308.12950}.

\bibitem[{Sadat et~al.(2023)Sadat, Zhou, Lange, Araki, Gundroo, Wang, Menon, Parvez, and Feng}]{sadat-etal-2023-delucionqa}
Mobashir Sadat, Zhengyu Zhou, Lukas Lange, Jun Araki, Arsalan Gundroo, Bingqing Wang, Rakesh Menon, Md~Parvez, and Zhe Feng. 2023.
\newblock \href {https://doi.org/10.18653/v1/2023.findings-emnlp.59} {{D}elucion{QA}: Detecting hallucinations in domain-specific question answering}.
\newblock In \emph{Findings of the Association for Computational Linguistics: EMNLP 2023}, pages 822--835, Singapore. Association for Computational Linguistics.

\bibitem[{Shi et~al.(2024)Shi, Wang, Wan, and Gu}]{shi2024code-mgdebug}
Yuling Shi, Songsong Wang, Chengcheng Wan, and Xiaodong Gu. 2024.
\newblock From code to correctness: Closing the last mile of code generation with hierarchical debugging.
\newblock \emph{arXiv preprint arXiv:2410.01215}.

\bibitem[{Shinn et~al.(2023)Shinn, Cassano, Gopinath, Narasimhan, and Yao}]{shinn2023reflexion}
Noah Shinn, Federico Cassano, Ashwin Gopinath, Karthik~R Narasimhan, and Shunyu Yao. 2023.
\newblock Reflexion: Language agents with verbal reinforcement learning.
\newblock In \emph{Thirty-seventh Conference on Neural Information Processing Systems}.

\bibitem[{Shum et~al.(2023)Shum, Diao, and Zhang}]{shum-etal-2023-automatic}
Kashun Shum, Shizhe Diao, and Tong Zhang. 2023.
\newblock \href {https://doi.org/10.18653/v1/2023.findings-emnlp.811} {Automatic prompt augmentation and selection with chain-of-thought from labeled data}.
\newblock In \emph{Findings of the Association for Computational Linguistics: EMNLP 2023}, pages 12113--12139, Singapore. Association for Computational Linguistics.

\bibitem[{Touvron et~al.(2023)Touvron, Martin, Stone, Albert, Almahairi, Babaei, Bashlykov, Batra, Bhargava, Bhosale et~al.}]{touvron2023llama}
Hugo Touvron, Louis Martin, Kevin Stone, Peter Albert, Amjad Almahairi, Yasmine Babaei, Nikolay Bashlykov, Soumya Batra, Prajjwal Bhargava, Shruti Bhosale, et~al. 2023.
\newblock \href {https://arxiv.org/pdf/2307.09288} {Llama 2: Open foundation and fine-tuned chat models}.
\newblock \emph{arXiv preprint arXiv:2307.09288}.

\bibitem[{Wang et~al.(2023{\natexlab{a}})Wang, Wei, Schuurmans, Le, Chi, Narang, Chowdhery, and Zhou}]{wang2023selfconsistency}
Xuezhi Wang, Jason Wei, Dale Schuurmans, Quoc Le, Ed~Chi, Sharan Narang, Aakanksha Chowdhery, and Denny Zhou. 2023{\natexlab{a}}.
\newblock \href {http://arxiv.org/abs/2203.11171} {Self-consistency improves chain of thought reasoning in language models}.

\bibitem[{Wang et~al.(2021)Wang, Wang, Joty, and Hoi}]{wang2021codet5}
Yue Wang, Weishi Wang, Shafiq Joty, and Steven~CH Hoi. 2021.
\newblock Codet5: Identifier-aware unified pre-trained encoder-decoder models for code understanding and generation.
\newblock \emph{arXiv preprint arXiv:2109.00859}.

\bibitem[{Wang et~al.(2023{\natexlab{b}})Wang, Araki, Jiang, Parvez, and Neubig}]{filco}
Zhiruo Wang, Jun Araki, Zhengbao Jiang, Md~Rizwan Parvez, and Graham Neubig. 2023{\natexlab{b}}.
\newblock Learning to filter context for retrieval-augmented generation.
\newblock \emph{arXiv preprint arXiv:2311.08377}.

\bibitem[{Wei et~al.(2022{\natexlab{a}})Wei, Wang, Schuurmans, Bosma, Xia, Chi, Le, Zhou et~al.}]{CoT}
Jason Wei, Xuezhi Wang, Dale Schuurmans, Maarten Bosma, Fei Xia, Ed~Chi, Quoc~V Le, Denny Zhou, et~al. 2022{\natexlab{a}}.
\newblock Chain-of-thought prompting elicits reasoning in large language models.
\newblock \emph{Advances in Neural Information Processing Systems}, 35:24824--24837.

\bibitem[{Wei et~al.(2022{\natexlab{b}})Wei, Wang, Schuurmans, Bosma, Xia, Chi, Le, Zhou et~al.}]{wei2022chain}
Jason Wei, Xuezhi Wang, Dale Schuurmans, Maarten Bosma, Fei Xia, Ed~Chi, Quoc~V Le, Denny Zhou, et~al. 2022{\natexlab{b}}.
\newblock Chain-of-thought prompting elicits reasoning in large language models.
\newblock \emph{Advances in Neural Information Processing Systems}, 35:24824--24837.

\bibitem[{Yasunaga et~al.(2023)Yasunaga, Chen, Li, Pasupat, Leskovec, Liang, Chi, and Zhou}]{yasunaga2023large}
Michihiro Yasunaga, Xinyun Chen, Yujia Li, Panupong Pasupat, Jure Leskovec, Percy Liang, Ed~H Chi, and Denny Zhou. 2023.
\newblock Large language models as analogical reasoners.
\newblock \emph{arXiv preprint arXiv:2310.01714}.

\bibitem[{Yin and Neubig(2017)}]{cmu_code_gen}
Pengcheng Yin and Graham Neubig. 2017.
\newblock \href {http://arxiv.org/abs/1704.01696} {A syntactic neural model for general-purpose code generation}.
\newblock \emph{CoRR}, abs/1704.01696.

\bibitem[{Yu et~al.(2019)Yu, Zhang, Er, Li, Xue, Pang, Lin, Tan, Shi, Li, Jiang, Yasunaga, Shim, Chen, Fabbri, Li, Chen, Zhang, Dixit, Zhang, Xiong, Socher, Lasecki, and Radev}]{yu-etal-2019-cosql}
Tao Yu, Rui Zhang, Heyang Er, Suyi Li, Eric Xue, Bo~Pang, Xi~Victoria Lin, Yi~Chern Tan, Tianze Shi, Zihan Li, Youxuan Jiang, Michihiro Yasunaga, Sungrok Shim, Tao Chen, Alexander Fabbri, Zifan Li, Luyao Chen, Yuwen Zhang, Shreya Dixit, Vincent Zhang, Caiming Xiong, Richard Socher, Walter Lasecki, and Dragomir Radev. 2019.
\newblock \href {https://doi.org/10.18653/v1/D19-1204} {{C}o{SQL}: A conversational text-to-{SQL} challenge towards cross-domain natural language interfaces to databases}.
\newblock In \emph{Proceedings of the 2019 Conference on Empirical Methods in Natural Language Processing and the 9th International Joint Conference on Natural Language Processing (EMNLP-IJCNLP)}, pages 1962--1979, Hong Kong, China. Association for Computational Linguistics.

\bibitem[{Zhang et~al.(2024)Zhang, Li, Li, Shi, and Jin}]{zhang-etal-2024-codeagent}
Kechi Zhang, Jia Li, Ge~Li, Xianjie Shi, and Zhi Jin. 2024.
\newblock \href {https://doi.org/10.18653/v1/2024.acl-long.737} {{C}ode{A}gent: Enhancing code generation with tool-integrated agent systems for real-world repo-level coding challenges}.
\newblock In \emph{Proceedings of the 62nd Annual Meeting of the Association for Computational Linguistics (Volume 1: Long Papers)}, pages 13643--13658, Bangkok, Thailand. Association for Computational Linguistics.

\bibitem[{Zhang et~al.(2022)Zhang, Zhang, Li, and Smola}]{zhang2022automatic}
Zhuosheng Zhang, Aston Zhang, Mu~Li, and Alex Smola. 2022.
\newblock Automatic chain of thought prompting in large language models.
\newblock \emph{arXiv preprint arXiv:2210.03493}.

\bibitem[{Zhao et~al.(2023)Zhao, Zhou, Li, Tang, Wang, Hou, Min, Zhang, Zhang, Dong, Du, Yang, Chen, Chen, Jiang, Ren, Li, Tang, Liu, Liu, Nie, and Wen}]{LLMSurvey}
Wayne~Xin Zhao, Kun Zhou, Junyi Li, Tianyi Tang, Xiaolei Wang, Yupeng Hou, Yingqian Min, Beichen Zhang, Junjie Zhang, Zican Dong, Yifan Du, Chen Yang, Yushuo Chen, Zhipeng Chen, Jinhao Jiang, Ruiyang Ren, Yifan Li, Xinyu Tang, Zikang Liu, Peiyu Liu, Jian-Yun Nie, and Ji-Rong Wen. 2023.
\newblock \href {http://arxiv.org/abs/2303.18223} {A survey of large language models}.
\newblock \emph{arXiv preprint arXiv:2303.18223}.

\bibitem[{Zhong et~al.(2024)Zhong, Wang, and Shang}]{zhong-etal-2024-debug}
Li~Zhong, Zilong Wang, and Jingbo Shang. 2024.
\newblock \href {https://doi.org/10.18653/v1/2024.findings-acl.49} {Debug like a human: A large language model debugger via verifying runtime execution step by step}.
\newblock In \emph{Findings of the Association for Computational Linguistics: ACL 2024}, pages 851--870, Bangkok, Thailand. Association for Computational Linguistics.

\bibitem[{Zhou et~al.(2023)Zhou, Yan, Shlapentokh-Rothman, Wang, and Wang}]{zhou2023lats}
Andy Zhou, Kai Yan, Michal Shlapentokh-Rothman, Haohan Wang, and Yu-Xiong Wang. 2023.
\newblock Language agent tree search unifies reasoning acting and planning in language models.
\newblock \emph{arXiv preprint arXiv:2310.04406}.

\end{thebibliography}

\newpage

\vspace*{\textheight}
\newpage

\noindent {\LARGE \textbf{Appendix}} \\

\section{Algorithm of \ourapproach}
\label{alg:codesim}

Algorithm~1 shows the pseudo-code of our prompting technique.

\begin{algorithm}[h]
\small
\caption{\tool}
\begin{algorithmic}[1]

\State $p \gets$ maximum number of planning steps
\State $d \gets$ maximum number of debugging steps \\

\For{$i \gets 1$ to $p$}
    \State \# Start of Planning Agent
    \State $plan \gets$ GeneratePlan($problem$)
    \State $feedback \gets$ ValidatePlan($problem$, $plan$)
    \If{$feedback$ is negative}
        \State $plan \gets$ RefinePlan($problem$, $plan$, $feedback$)
    \EndIf
    \State \# End of Planning Agent
    \\
    \State \# Start of Coding Agent
    \State $code \gets$ GenerateCode($problem$, $plan$) 
    
    \State $passed,\space log \gets$ test($code$, $sample\_io$) 
    \If{$passed$}
        \State Return $code$
    \Else
        \State \# Start of Debugging Agent
        \For{$j \gets 1$ to $d$}
            \State $code \gets$ DebugCode(\\
                \quad \quad \quad \quad \quad $problem$,\\
                \quad \quad \quad \quad \quad $plan$, \\
                \quad \quad \quad \quad \quad $code$, \\
                \quad \quad \quad \quad \quad $log$ \\
            \quad \quad \quad \quad) \\ 
            \State $passed,\space log \gets$ test($code$, $sample\_io$)
            \If{$passed$}
                \State Return $code$
            \EndIf
        \EndFor
        \State \# End of Debugging Agent
    \EndIf
    \State \# End of Coding Agent
    \\
\EndFor
\State Return $code$
\end{algorithmic}
\end{algorithm}

\section{Exclusion of AgentCoder}
\label{app:elimination-of-agentcoder}
We have not included AgentCoder \cite{huang2023agentcoder} in our comparison due to reproducibility issues which undoubtedly plays a critical role in fair comparison as indicted in \citet{laskar-etal-2024-systematic}, as we were unable to replicate their results. In our attempts to reproduce their work on the HumanEval benchmark using ChatGPT, we achieved $56.7$\% accuracy after four iterations, consuming $11.9$ million tokens. When using GPT-4, we attained only $17.7$\% accuracy after two iterations, with $10.4$ million tokens consumed. The token consumption in both cases is significantly higher compared to MapCoder ($1.7$ million tokens with ChatGPT and $2.1$ million with GPT-4) and \toolnospace($0.89$ million tokens in ChatGPT and $0.85$ million in GPT-4). These two experiments resulted in a cost of approximately \$$500$ USD, but we were still unable to come close to AgentCoder's reported claims of $79.9$\% accuracy with ChatGPT and $96.3$\% with GPT-4.

Furthermore, we found unaddressed issues on their GitHub page (\href{https://github.com/huangd1999/AgentCoder/issues/8}{link}) related to reproducibility. Additionally, for the MBPP dataset, they used all test cases as public test cases (\href{https://github.com/huangd1999/AgentCoder/issues/3}{link}), which deviates from standard practices. As a result, we did not consider those results in our comparison either.

\section{Details Promptings of \tool}
\label{app:prompts}

\begin{figure*}[h]
    \centering
    \includegraphics[width=0.90\textwidth]{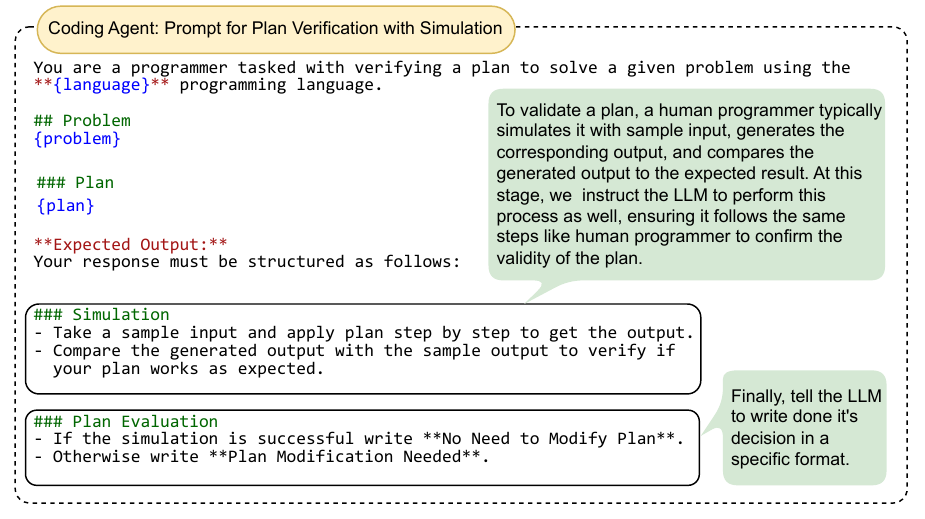}
    \caption{\emph{Planning Agent}: Prompt for Plan Verification with the help of Simulation.}
    \label{prompt:plan-verification}
\end{figure*}

\begin{figure*}[h]
    \centering
    \includegraphics[width=0.90\textwidth]{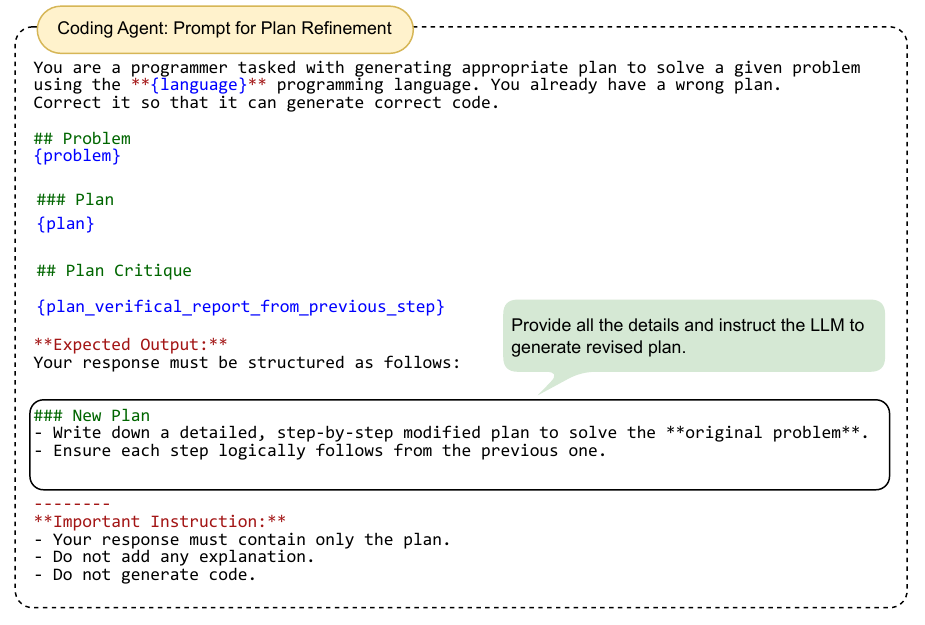}
    \caption{\emph{Planning Agent}: Prompt for Plan Refinement.}
    \label{prompt:plan-refinement}
\end{figure*}

The \emph{Planning Agent} interacts with the LLM three times to generate a plan. In the first API call, it instructs the LLM to comprehend the problem, generate an example problem, recommend a suitable algorithm, and finally produce the plan (Figure \ref{prompt:plan-generation}). In the second API call, the LLM is instructed to verify the plan through simulation (Figure \ref{prompt:plan-verification}). If the plan is satisfactory, it is returned by the agent. Otherwise, the LLM is called again to refine the plan based on the feedback from the simulation (Figure \ref{prompt:plan-refinement}).

\begin{figure*}[h]
    \centering
    \includegraphics[width=0.90\textwidth]{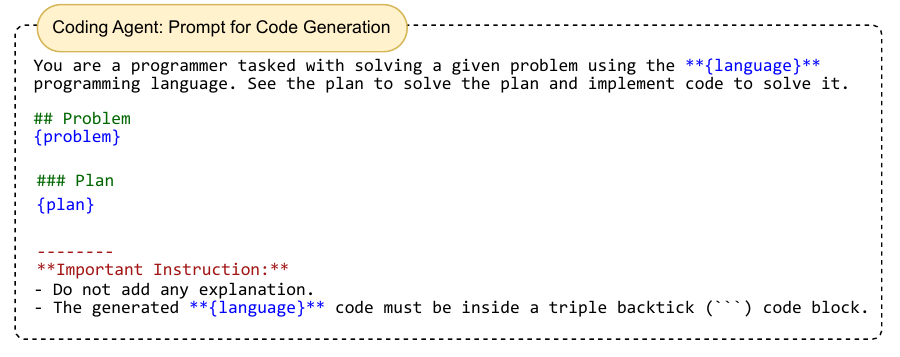}
    \caption{\emph{Coding Agent}: Prompt for Code Generation.}
    \label{prompt:code-generation}
\end{figure*}

The next step involves the \emph{Coding Agent}, which receives the plan from the \emph{Planning Agent} and uses the prompt outlined in Figure \ref{prompt:code-generation} to generate code.

\begin{figure*}[h]
    \centering
    \includegraphics[width=0.90\textwidth]{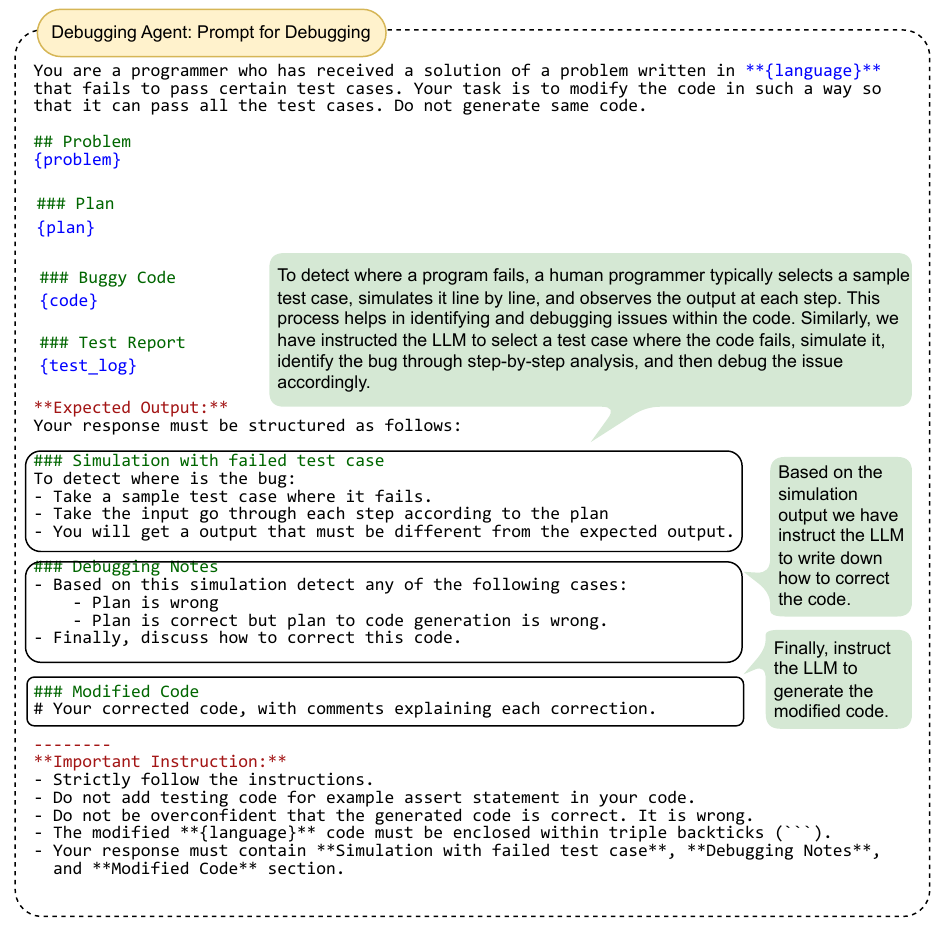}
    \caption{\emph{Debugging Agent}: Prompt for Debugging.}
    \label{prompt:debugging}    
\end{figure*}

If the code fails to pass the sample input/output, \tool activates its final agent, the \emph{Debugging Agent}, using the prompt shown in Figure \ref{prompt:debugging}.

These figures also include the rationale behind the inclusion of each sentence in the prompt.

\section{Example Problem}
We present a complete example of problem solving using \tool below:


\begin{figure*}[h]
    \centering
    \includegraphics[width=0.90\textwidth]{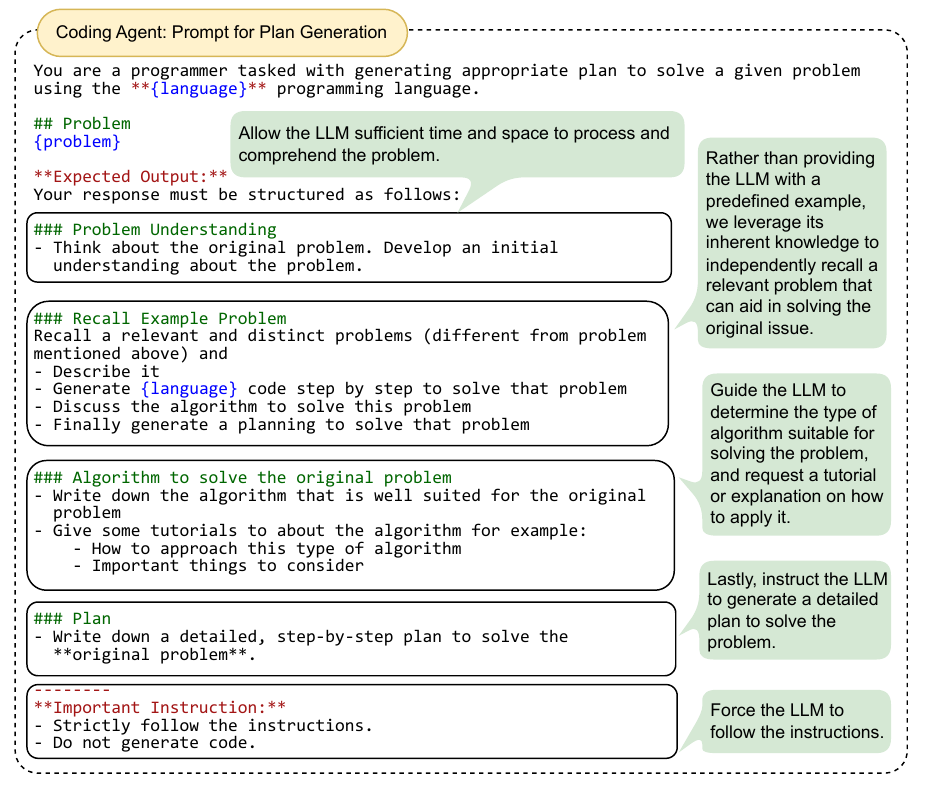}
    \caption{\emph{Planning Agent}: Prompt for Plan Generation. }
    \label{prompt:plan-generation}
\end{figure*} 

\onecolumn

\begin{tcolorbox}[breakable]
\label{app:example-problem}
    \vspace{2mm}
\textbf{An Example from HumanEval dataset for demonstrating how \tool works}
\hrule 
\vspace{5mm}
Input for Planning: 1

You are a programmer tasked with generating appropriate plan to solve a given problem using the \*\*Python3\*\* programming language.\\
 \\
\#\# Problem

\begin{verbatim}
def generate_integers(a, b):
    """
    Given two positive integers a and b, return the even digits between a
    and b, in ascending order.

    For example:
    generate_integers(2, 8) => [2, 4, 6, 8]
    generate_integers(8, 2) => [2, 4, 6, 8]
    generate_integers(10, 14) => []
    """
\end{verbatim}

\*\*Expected Output:\*\*\\
 \\
Your response must be structured as follows:\\

\#\#\# Problem Understanding\\
 \\
- Think about the original problem. Develop an initial understanding about the problem.\\

\#\#\# Recall Example Problem\\
 \\
Recall a relevant and distinct problems (different from problem mentioned above) and\\
- Describe it\\
- Generate Python3 code step by step to solve that problem\\
- Discuss the algorithm to solve this problem\\
- Finally generate a planning to solve that problem\\

\#\#\# Algorithm to solve the original problem\\
 \\
- Write down the algorithm that is well suited for the original problem\\
- Give some tutorials to about the algorithm for example:\\
    - How to approach this type of algorithm\\
    - Important things to consider\\

\#\#\# Plan\\
 \\
- Write down a detailed, step-by-step plan to solve the \*\*original problem\*\*.\\

--------\\
\*\*Important Instruction:\*\*\\
- Strictly follow the instructions.\\
- Do not generate code.\\

\vspace{0.75cm} 
\hrule
\vspace{0.75cm}

Response from Planning: 1\\

\#\#\# Problem Understanding

The problem requires generating a list of even integers between two given positive integers, $a$ and $b$, inclusive. The list should be in ascending order, regardless of whether $a$ is less than or greater than $b$. If there are no even integers in the range, the function should return an empty list. \\

\#\#\# Recall Example Problem

\*\*Example Problem:\*\*\\

Find all odd numbers between two given integers $x$ and $y$, inclusive, and return them in descending order.\\

\*\*Step-by-step Solution:\*\*\\

1. Determine the range between $x$ and $y$.\\
2. Identify all odd numbers within this range.\\
3. Sort the odd numbers in descending order.\\
4. Return the sorted list.\\

\*\*Algorithm:\*\*\\
 \\
1. Check if $x$ is greater than $y$. If so, swap them to ensure the range is correct.\\
2. Iterate through the range from $x$ to $y$.\\
3. For each number, check if it is odd (i.e., $number \% 2 != 0$).\\
4. Collect all odd numbers in a list.\\
5. Sort the list in descending order.\\
6. Return the sorted list.\\

\*\*Planning:\*\*\\
 \\
- Define a function $find\_odd\_numbers(x, y)$.\\
- Use a conditional to swap $x$ and $y$ if $x > y$.\\
- Use a loop to iterate through the range from $x$ to $y$.\\
- Check each number for oddness and add it to a list if it is odd.\\
- Sort the list in descending order.\\
- Return the list.\\

\#\#\# Algorithm to Solve the Original Problem\\
 \\
1. \*\*Determine the Range:\*\*\\
   - Identify the start and end of the range by taking the minimum and maximum of $a$ and $b$.\\

2. \*\*Iterate Through the Range:\*\*\\
   - Loop through each number in the range from the minimum to the maximum of $a$ and $b$.\\

3. \*\*Check for Even Numbers:\*\*\\
   - For each number, check if it is even by using the condition $number \% 2 == 0$.\\

4. \*\*Collect Even Numbers:\*\*\\
   - Add each even number to a list.\\

5. \*\*Return the List:\*\*\\
   - Since the numbers are collected in ascending order, return the list directly.\\
 \\
\*\*Tutorials:\*\*\\
 \\
- \*\*Approach:\*\*\\
  - Always ensure the range is correctly defined, especially when the order of inputs can vary.\\
  - Use modulus operation to check for evenness or oddness.\\
 \\
- \*\*Important Considerations:\*\*\\
  - Handle edge cases where there might be no even numbers in the range.\\
  - Ensure the function is robust to handle any positive integer inputs.\\

\#\#\# Plan \\
 \\
1. \*\*Define the Function:\*\* \\
   - Create a function $generate\_integers(a, b)$. \\

2. \*\*Determine the Range:\*\* \\
   - Use $min(a, b)$ and $max(a, b)$ to define the start and end of the range. \\

3. \*\*Initialize an Empty List:\*\* \\
   - Prepare an empty list to store even numbers. \\

4. \*\*Loop Through the Range:\*\* \\
   - Use a $for$ loop to iterate from the start to the end of the range. \\

5. \*\*Check for Evenness:\*\* \\
   - Within the loop, use $if number \% 2 == 0$ to check if a number is even. \\

6. \*\*Collect Even Numbers:\*\* \\
   - Append even numbers to the list. \\

7. \*\*Return the Result:\*\* \\
   - Return the list of even numbers, which will be in ascending order. \\

By following this plan, the function will correctly generate a list of even integers between the given numbers $a$ and $b$. \\
 \\

\vspace{0.5cm} \hrule   \vspace{0.5cm}
Input for Simulation: 1

You are a programmer tasked with verifying a plan to solve a given problem using the \*\*Python3\*\* programming language. \\

\#\# Problem: \\
\begin{verbatim}
def generate_integers(a, b):
    """
    Given two positive integers a and b, return the even digits between a
    and b, in ascending order.

    For example:
    generate_integers(2, 8) => [2, 4, 6, 8]
    generate_integers(8, 2) => [2, 4, 6, 8]
    generate_integers(10, 14) => []
    """
\end{verbatim}

\#\#\# Plan \\

1. \*\*Define the Function:\*\* \\
   - Create a function $generate\_integers(a, b)$. \\

2. \*\*Determine the Range:\*\* \\
   - Use $min(a, b)$ and $max(a, b)$ to define the start and end of the range.v

3. \*\*Initialize an Empty List:\*\* \\
   - Prepare an empty list to store even numbers. \\

4. \*\*Loop Through the Range:\*\* \\
   - Use a $for$ loop to iterate from the start to the end of the range. \\

5. \*\*Check for Evenness:\*\* \\
   - Within the loop, use $if \space number \% 2 == 0$ to check if a number is even. \\

6. \*\*Collect Even Numbers:\*\* \\
   - Append even numbers to the list. \\

7. \*\*Return the Result:\*\* \\
   - Return the list of even numbers, which will be in ascending order. \\

By following this plan, the function will correctly generate a list of even integers between the given numbers $a$ and $b$. \\

\*\*Expected Output:\*\* \\
 \\
Your response must be structured as follows: \\

\#\#\# Simulation \\
 \\
- Take a sample input and apply plan step by step to get the output. \\
- Compare the generated output with the sample output to verify if your plan works as expected. \\

\#\#\# Plan Evaluation \\
 \\
- If the simulation is successful write \*\*No Need to Modify Plan\*\*. \\
- Otherwise write \*\*Plan Modification Needed\*\*. \\

\vspace{0.5cm} \hrule   \vspace{0.5cm}
Response from Simulation: 1

\#\#\# Simulation \\
 \\
Let's apply the plan step by step using the sample input $generate\_integers(2, 8)$. \\

1. \*\*Define the Function:\*\* \\
   - We define the function $generate\_integers(a, b)$. \\

2. \*\*Determine the Range:\*\* \\
   - Calculate $start = min(2, 8) = 2$ and $end = max(2, 8) = 8$. \\

3. \*\*Initialize an Empty List:\*\* \\
   - Create an empty list $even_numbers = []$. \\

4. \*\*Loop Through the Range:\*\* \\
   - Iterate over the range from $start$ to $end$ (inclusive). \\

5. \*\*Check for Evenness:\*\* \\
   - For each number in the range, check if it is even using $if \space number \% 2 == 0$. \\

6. \*\*Collect Even Numbers:\*\* \\
   - Append even numbers to the list $even_numbers$. \\

7. \*\*Return the Result:\*\* \\
   - Return the list $even_numbers$. \\

Applying these steps: \\

- Range: 2 to 8 \\
- Iteration: \\
  - 2 is even, append to list: $[2]$ \\
  - 3 is not even \\
  - 4 is even, append to list: $[2, 4]$ \\
  - 5 is not even \\
  - 6 is even, append to list: $[2, 4, 6]$ \\
  - 7 is not even \\
  - 8 is even, append to list: $[2, 4, 6, 8]$ \\

The output is $[2, 4, 6, 8]$, which matches the expected output. \\

\#\#\# Plan Evaluation \\
 \\
\*\*No Need to Modify Plan\*\* \\

The plan works as expected for the given problem. The function correctly generates a list of even integers between the given numbers $a$ and $b$, regardless of their order. \\

\vspace{0.5cm} \hrule   \vspace{0.5cm}
Input for final code generation: \\
 \\

You are a programmer tasked with solving a given problem using the \*\*Python3\*\* programming language. See the plan to solve the plan and implement code to solve it. \\

\#\# Problem: \\
\begin{verbatim}
def generate_integers(a, b):
    """
    Given two positive integers a and b, return the even digits between a
    and b, in ascending order.

    For example:
    generate_integers(2, 8) => [2, 4, 6, 8]
    generate_integers(8, 2) => [2, 4, 6, 8]
    generate_integers(10, 14) => []
    """
\end{verbatim}

\#\#\# Plan \\
 \\
1. \*\*Define the Function:\*\* \\
   - Create a function $generate\_integers(a, b)$. \\

2. \*\*Determine the Range:\*\* \\
   - Use $min(a, b)$ and $max(a, b)$ to define the start and end of the range. \\

3. \*\*Initialize an Empty List:\*\* \\
   - Prepare an empty list to store even numbers. \\

4. \*\*Loop Through the Range:\*\* \\
   - Use a $for$ loop to iterate from the start to the end of the range. \\

5. \*\*Check for Evenness:\*\* \\
   - Within the loop, use $if  \space number \% 2 == 0$ to check if a number is even. \\

6. \*\*Collect Even Numbers:\*\* \\
   - Append even numbers to the list. \\

7. \*\*Return the Result:\*\* \\
   - Return the list of even numbers, which will be in ascending order. \\
 \\
By following this plan, the function will correctly generate a list of even integers between the given numbers $a$ and $b$. \\

-------- \\
\*\*Important Instructions:\*\* \\
- Do not add any explanation. \\
- The generated \*\*Python3\*\* code must be inside a triple backtick (\$\$\$) code block. \\ \\

\vspace{0.5cm} \hrule   \vspace{0.5cm}
Response from final code generation: \\
 \\

\begin{verbatim}
def generate_integers(a, b):
    start = min(a, b)
    end = max(a, b)
    even_numbers = []
    for number in range(start, end + 1):
        if number % 2 == 0:
            even_numbers.append(number)
    return even_numbers
\end{verbatim}

\vspace{0.5cm} \hrule   \vspace{0.5cm}
Input for Improving code: 1, 1 \\
 \\
You are a programmer who has received a solution of a problem written in \*\*Python3\*\* that fails to pass certain test cases. Your task is to modify the code in such a way so that it can pass all the test cases. Do not generate same code. \\

\#\# Problem: \\
\begin{verbatim}
def generate_integers(a, b):
    """
    Given two positive integers a and b, return the even digits between a
    and b, in ascending order.

    For example:
    generate_integers(2, 8) => [2, 4, 6, 8]
    generate_integers(8, 2) => [2, 4, 6, 8]
    generate_integers(10, 14) => []
    """
\end{verbatim}
\#\#\# Plan \\
 \\
1. \*\*Define the Function:\*\* \\
   - Create a function $generate\_integers(a, b)$. \\

2. \*\*Determine the Range:\*\* \\
   - Use $min(a, b)$ and $max(a, b)$ to define the start and end of the range. \\

3. \*\*Initialize an Empty List:\*\* \\
   - Prepare an empty list to store even numbers. \\

4. \*\*Loop Through the Range:\*\* \\
   - Use a $for$ loop to iterate from the start to the end of the range. \\

5. \*\*Check for Evenness:\*\* \\
   - Within the loop, use $if \space number \% 2 == 0$ to check if a number is even. \\

6. \*\*Collect Even Numbers:\*\* \\
   - Append even numbers to the list. \\

7. \*\*Return the Result:\*\* \\
   - Return the list of even numbers, which will be in ascending order. \\

By following this plan, the function will correctly generate a list of even integers between the given numbers $a$ and $b$. \\

\#\#\# Buggy Code \\
 \\
\begin{verbatim}
def generate_integers(a, b):
    start = min(a, b)
    end = max(a, b)
    even_numbers = []
    for number in range(start, end + 1):
        if number % 2 == 0:
            even_numbers.append(number)
    return even_numbers
\end{verbatim}

\#\#\# Test Report \\
 \\
Test Cases where the generated code failed to generate the expected output: \\
$assert generate_integers(10, 14) == []$ \\

\*\*Expected Output:\*\* \\
 \\
Your response must be structured as follows: \\

\#\#\# Simulation with failed test case \\
To detect where is the bug: \\
    - Take a sample test case where it fails. \\
    - Take the input go through each step according to the plan \\
    - You will get a output that must be different from the expected output.  \\

\#\#\# Debugging Notes \\
Based on this simulation detect any of the following cases: \\
    - Plan is wrong \\
    - Plan is correct but plan to code generation is wrong. \\

- Finally, discuss how to correct this code. \\

\#\#\# Modified Code \\
 \\
\texttt{```}Python3 \\
\# Your corrected code, with comments explaining each correction. \\
\texttt{```} \\

-------- \\
\*\*Important Instructions:\*\* \\
- Strictly follow the instructions. \\
- Do not add testing code for example assert statement in your code. \\
- Do not be overconfident that the generated code is correct. It is wrong. \\
- The modified \*\*Python3\*\* code must be enclosed within triple backticks (\texttt{```}). \\
- Your response must contain \*\*Simulation with failed test case\*\*, \*\*Debugging Notes\*\*, and \*\*Modified Code\*\* section. \\

\vspace{0.5cm} \hrule   \vspace{0.5cm}
Response from Improving code: 1, 1 \\
 \\

\#\#\# Simulation with failed test case \\
 \\
Let's take the test case $generate\_integers(10, 14)$:v

1. \*\*Determine the Range:\*\* \\
   - $start = min(10, 14) = 10$ \\
   - $end = max(10, 14) = 14$ \\

2. \*\*Loop Through the Range:\*\* \\
   - The loop iterates over $10, 11, 12, 13, 14$. \\

3. \*\*Check for Evenness:\*\* \\
   - $10 \% 2 == 0$ (even, added to list) \\
   - $11 \% 2 != 0$ (odd, not added) \\
   - $12 \% 2 == 0$ (even, added to list) \\
   - $13 \% 2 != 0$ (odd, not added) \\
   - $14 \% 2 == 0$ (even, added to list) \\

4. \*\*Collect Even Numbers:\*\* \\
   - The list becomes $[10, 12, 14]$. \\

5. \*\*Return the Result:\*\* \\
   - The function returns $[10, 12, 14]$. \\

\*\*Expected Output:\*\* $[]$ \\

\#\#\# Debugging Notes \\
 \\
- \*\*Plan is correct\*\*: The plan correctly identifies the need to find even numbers between $a$ and $b$. \\
- \*\*Plan to code generation is wrong\*\*: The code does not account for the requirement that the even digits themselves must be between $a$ and $b$. It incorrectly includes numbers like $10, 12, 14$ when the digits themselves (1, 2, 4) are not between $10$ and $14$. \\

\#\#\# Modified Code \\
\begin{verbatim}
def generate_integers(a, b):
    start = min(a, b)
    end = max(a, b)
    even_numbers = []
    for number in range(start, end + 1):
        # Check each digit in the number
        for digit in str(number):
            if int(digit) % 2 == 0 and start <= int(digit) <= end:
                even_numbers.append(int(digit))
    # Remove duplicates and sort the result
    return sorted(set(even_numbers))
\end{verbatim}

\*\*Explanation of Corrections:\*\* \\
 \\
- \*\*Digit Check\*\*: The code now checks each digit of the numbers in the range to see if it is even and within the specified range. \\
- \*\*Set for Uniqueness\*\*: A set is used to ensure that each even digit is only included once. \\
- \*\*Sorting\*\*: The final list is sorted to ensure ascending order. \\

\end{tcolorbox}
\twocolumn  

\end{document}